\documentclass{article}

\usepackage[final]{neurips_2021}

\usepackage[utf8]{inputenc} 
\usepackage[T1]{fontenc}    
\usepackage{hyperref}       
\usepackage{url}            
\usepackage{booktabs}       
\usepackage{amsfonts}       
\usepackage{nicefrac}       
\usepackage{microtype}      

\usepackage{nicefrac}       
\usepackage{amsmath}        
\usepackage{bm}             
\usepackage{dsfont}         
\usepackage{graphicx}
\usepackage{sidecap}
\usepackage{caption}
\usepackage{multirow}
\usepackage{xcolor}
\usepackage{tcolorbox}
\usepackage{multicol}
\usepackage{amssymb}

\usepackage{enumitem}
\usepackage{pifont}

\usepackage{amsmath,amsfonts,bm}

\def\eqref#1{equation~\ref{#1}}

\def\1{\bm{1}}

\DeclareMathAlphabet{\mathsfit}{\encodingdefault}{\sfdefault}{m}{sl}
\SetMathAlphabet{\mathsfit}{bold}{\encodingdefault}{\sfdefault}{bx}{n}

\newcommand{\LL}{\mathcal{L}}

\newcommand{\eqdef}{\overset{\mathrm{{\small def}}}{=}}

\title{Local plasticity rules can learn deep representations using self-supervised contrastive predictions}

\makeatletter
\let\newtitle\@title
\makeatother

\author{%
  Bernd Illing
  \And
  Jean Ventura
  \AND
  Guillaume Bellec\thanks{shared last author}
  \And
  Wulfram Gerstner\footnotemark[1]
  \and
  ~\\
  \texttt{\{firstname.lastname\}@epfl.ch}\\
  ~\\
  Department of Computer Science \& Department of Life Sciences\\
  École Polytechnique Fédérale de Lausanne\\
  1015 Switzerland
}

\begin{document}

\maketitle

\begin{abstract}
    Learning in the brain is poorly understood and learning rules that respect biological constraints, yet yield deep hierarchical representations, are still unknown. 
    Here, we propose a learning rule that takes inspiration from neuroscience and recent advances in self-supervised deep learning. 
    Learning minimizes a simple layer-specific loss function and does not need to back-propagate error signals within or between layers. 
    Instead, weight updates follow a local, Hebbian, learning rule that only depends on pre- and post-synaptic neuronal activity, predictive dendritic input and widely broadcasted modulation factors which are identical for large groups of neurons. 
    The learning rule applies contrastive predictive learning to a causal, biological setting using saccades (i.e. rapid shifts in gaze direction).
    We find that networks trained with this self-supervised and local rule build deep hierarchical representations of images, speech and video. 
\end{abstract}

\section{Introduction}

Synaptic connection strengths in the brain are thought to change according to `Hebbian' plasticity rules \citep{Hebb1949}. 
Such rules are local and depend only on the recent state of the pre- and post-synaptic neurons \citep{sjostrom2001rate, caporale2008spike, markram2011history}, potentially modulated by a third factor related to reward, attention or other high-level signals \citep{Kusmierz2017, gerstner2018eligibility}.
Therefore, one appealing hypothesis is that {\it representation learning in sensory cortices emerges from local and unsupervised plasticity rules}. 

Following a common definition in the field \citep{Fukushima1988, Riesenhuber1999, lecun2012learning, lillicrap2020backpropagation}, a hierarchical representation (i) builds higher-level features out of lower-level ones, and (ii) provides more useful features in higher layers.
Now there seems to be a substantial gap between the rich hierarchical representations observed in the cortex and the representations emerging from local plasticity rules implementing principal/independent component analysis \citep{Oja1982, Hyvarinen1998}, sparse coding \citep{Olshausen1997, rozell2008sparse} or slow-feature analysis \citep{Foldiak1991, Wiskott2002,Sprekeler2007}.
Hebbian rules seem to struggle especially when `stacked', i.e. when asked to learn deep, hierarchical representations.

This performance gap is puzzling because there are learning rules, relying on back-propagation (BP), that {\it can} build hierarchical representations similar to those found in visual cortex \citep{yamins2014performance, zhuang2021unsupervised}. 
Although some progress towards biologically plausible implementations of back-propagation has been made \citep{lillicrap2016random, guerguiev2017towards, sacramento2018dendritic, payeur2021burst}, most models rely either on a neuron-specific error signal that needs to be transmitted by a separate error network \citep{crick1989, amit2019deep, Kunin2020}, or time-multiplexing feedforward and error signals \citep{lillicrap2020backpropagation, payeur2021burst}.
Algorithms like contrastive divergence \citep{hinton2002training}, contrastive Hebbian learning \citep{xie2003} or equilibrium propagation \citep{scellier2017equilibrium} use local activity exclusively to calculate updates, but they require to wait for convergence to an equilibrium which is not appropriate for online learning from quickly varying inputs.

The present paper demonstrates that deep representations can emerge from a local, biologically plausible and unsupervised learning rule, by integrating two important insights from neuroscience:
First, we focus on self-supervised learning from temporal data -- as opposed to supervised learning from labelled examples -- because this comes closest to natural data, perceived by real biological agents, and because the temporal structure of natural stimuli is a rich source of information.
In particular, we exploit the self-awareness of typical, self-generated changes of gaze direction (`\emph{saccades}') to distinguish input from a moving object during fixation from input arriving after a saccade towards a new object.
In our plasticity rule, a global factor modulates plasticity, depending on the presence or absence of such saccades.
Although we do not model the precise circuit that computes this global factor, we see it related to global, saccade-specific signals from motor areas, combined with surprise or prediction error, as in other models of synaptic plasticity \citep{angela2005uncertainty, nassar2012rational, heilbron2019confidence, liakoni2021surprise}.
Second, we notice that electrical signals stemming from segregated apical dendrites can modulate synaptic plasticity in biological neurons \citep{Kording2001, Major2013}, enabling context-dependent plasticity.

Algorithmically, our approach takes inspiration from deep self-supervised learning algorithms that seek to contrast, cluster or predict stimuli in the context of BP \citep{Oord2018,caron2018deep,zhuang2019local,Lowe2019}. 
Interestingly, \citet{Lowe2019} demonstrated that such methods even work if end-to-end BP is partially interrupted.
We build upon this body of work and suggest the {\it Contrastive, Local And Predictive Plasticity} (CLAPP) model which avoids BP completely, yet still builds hierarchical representations.\footnote{Our code is available at \url{https://github.com/EPFL-LCN/pub-illing2021-neurips}}



\section{Main goals and related work}\label{sec:related_work}

In this paper, we propose a local plasticity rule that learns deep representations.
To describe our model of synaptic plasticity, we represent a cortical area by the layer $l$ of a deep neural network. 
The neural activity of this layer at time $t$ is represented by the vector $\bm z^{t,l} = \rho(\bm a^{t,l})$, where $\rho$ is a non-linearity and $\bm a^{t,l} = \bm W^l \bm z^{t,l-1}$ is the vector of the respective summed inputs to the neurons through their basal dendrites $\bm W^l$ (the bias is absorbed into $\bm W^l$). 
To simplify notation, we write the pre-synaptic input as $\bm x^{t,l}=\bm z^{t,l-1}$ and we only specify the layer index $l$ when it is necessary.

\begin{figure}[t]
    \centering
    \includegraphics[width=\textwidth]{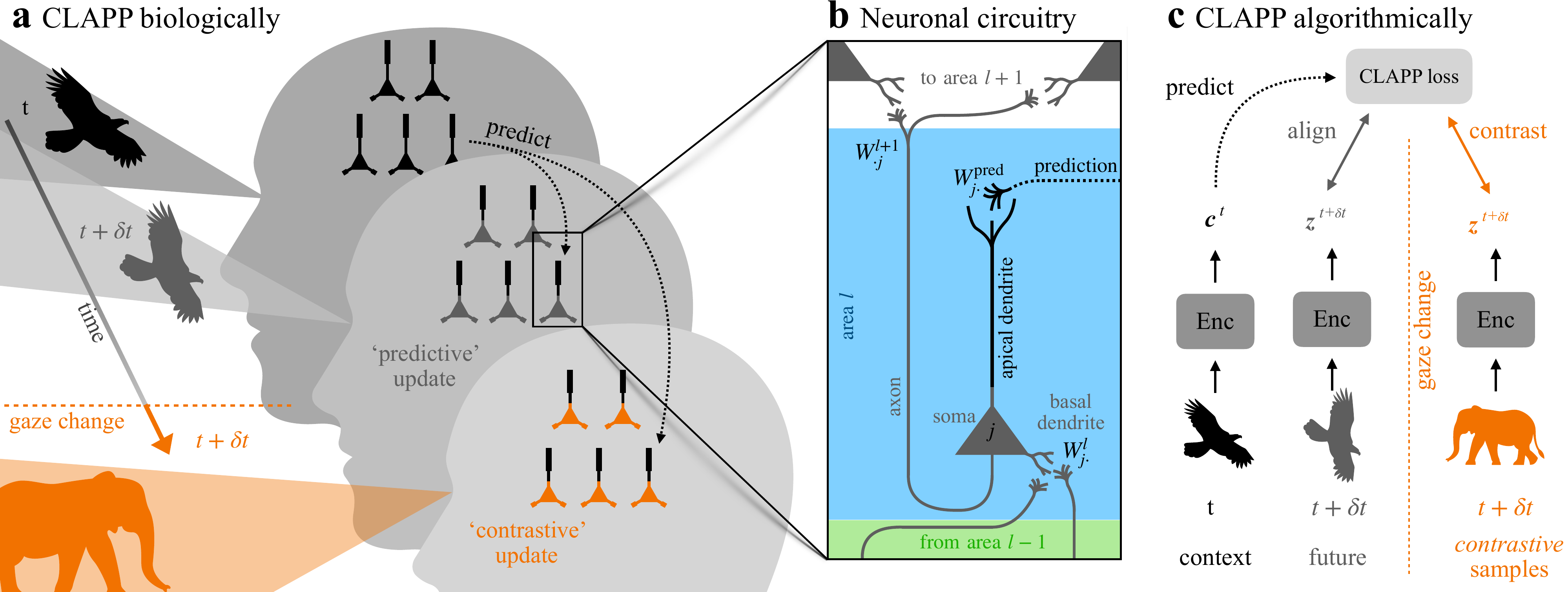}
    \caption{Contrastive, local and predictive plasticity (CLAPP). 
    \textbf{a} Perceiving a moving object (e.g. an eagle) at times $t$ and $t+\delta t$ leads to neural responses in the visual cortex. 
    After a gaze change (`saccade'), a different object (elephant) is seen.
    \textbf{b} (zoom) At each time step, pyramidal neurons integrate input activity at the basal dendrites (matrix $\bm W^{l}$ of feedforward weights) and pass on their response to downstream areas ($\bm W^{l+1}$). At any point in time, neurons predict future neural responses 
    through recurrent connections $\bm W^{\mathrm{pred}}$.
    These inputs target the apical dendrites and modulate ongoing synaptic plasticity through `predictive' updates. 
    Information about a saccade is transmitted by a broadcast signal triggered at the moment of saccade initiation, which leads to `contrastive' updates.
    As no external supervision or reward signals are provided, learning is self-supervised and local in time and space (`Hebbian').
    \textbf{c} Algorithmically, an encoder network (Enc) produces a `context' representation $\bm c^t$ at time $t$.
    Given $\bm c^t$, CLAPP tries to {\it predict} the encoding of the future input $\bm z^{t+\delta t}$. 
    In case of a gaze change between $t$ and $t+\delta t$, CLAPP seeks to keep the prediction as different as possible from the encoding of the upcoming {\it contrastive} sample.}
    \vspace{-0.5cm}
    \label{fig:fig0}
\end{figure}

Our plasticity rule exploits the fact that the temporal structure of natural inputs affects representation learning \citep{li2008unsupervised}.
Specifically, we consider a scenario where an agent first perceives a moving object at time $t$ (e.g. a flying eagle in \autoref{fig:fig0} a), and then spontaneously decides to change gaze direction towards another moving object at time $t+\delta t$ (e.g. {\it saccade} towards the elephant in \autoref{fig:fig0} a).
We further assume that the visual pathway is `self-aware' of saccades due to saccade-specific modulation of processing \citep{ross2001changes}.

In line with classical models of synaptic plasticity, we assume that weight changes follow biologically plausible, \emph{Hebbian}, learning rules \citep{Hebb1949, markram2011history} which are local in space and time:
updates $\Delta W^t_{ji}$ of a synapse, connecting neurons $i$ and $j$, can only depend on the current activity of the pre-synaptic and post-synaptic neurons at time $t$, or slightly earlier at time $t-\delta t$, and one or several widely broadcasted modulating factors \citep{urbanczik2009reinforcement, gerstner2018eligibility}.

Furthermore, we allow the activity of another neuron $k$ to influence the weight update $\Delta W_{ji}$, as long as there is an explicit connection $W_{jk}^\mathrm{pred}$ from $k$ to $j$.
The idea is to overcome the representational limitations of classical Hebbian learning by including dendritic inputs, which are thought to predict the future somatic activity \citep{Kording2001,urbanczik2014learning} and take part in the plasticity of the post-synaptic neuron \citep{Larkum1999, Dudman2007, Major2013}.
Hence we assume that each neuron $j$ in a layer $l$ may receive dendritic inputs $(\bm W^{\mathrm{pred}} \bm c^{t,l})_j$ coming either from the layer above ($\bm c^{t,l} = \bm z^{t,l+1}$) or from lateral connections in the same layer ($\bm c^{t,l} = \bm z^{t,l}$). 

For algorithmic reasons, that we detail in \autoref{sec:clapp_model}, we assume that the dendritic input $(\bm W^{\mathrm{pred}} \bm c^{t,l})_j$ influences the weight updates $\Delta W_{ji}$ of the post-synaptic neuron $j$, but not its activity $z_j^t$.
This assumption is justified by neuroscientific findings that the inputs to basal and apical dendrites affect the neural activity and plasticity in different ways \citep{Larkum1999, Dudman2007, Major2013, urbanczik2014learning}.
In general, we do not rule out influence of dendritic activity on somatic activity in later processing phases, but see this beyond the scope of the current work.

Given these insights from neuroscience, we gather the essential factors that influence synaptic plasticity in the following learning rule prototype:
\begin{eqnarray}
\Delta W_{ji} & \propto & \underbrace{~\mathrm{modulators}~}_{\substack{\text{broadcast factors}}} ~
\cdot\underbrace{(\bm W^{\mathrm{pred}} \bm c^{t_1})_j}_{\text{dendritic prediction}} ~
\cdot\underbrace{\mathrm{post}^{t_2}_j ~
\cdot \mathrm{pre}^{t_2}_i ~ 
}_{\text{local-activity}}~.
\label{eq:rule_prototype}
\end{eqnarray}
The modulating broadcast factors are the same for large groups of neurons, for example all neurons in the same area, or even all neurons in the whole network.
$\mathrm{post}^{t_2}_j$ and $\mathrm{pre}^{t_2}_i$ are functions of the pre- and post- synaptic activities. At this point, we do not specify the exact timing between $t_1$ and $t_2$, as this will be determined by our algorithm in \autoref{sec:gradient_analysis}.

\begin{figure}[t]
    \centering
    \includegraphics[width=1\textwidth]{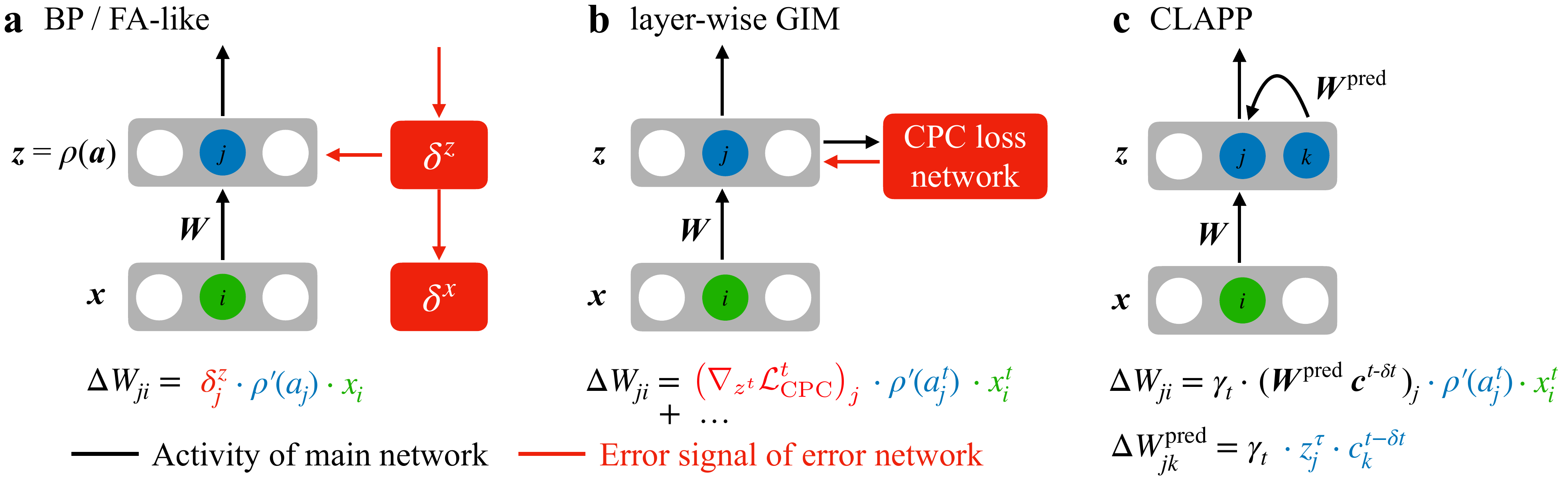}
    \caption{
            Comparison of weight updates 
            \textbf{a} Networks trained with back-propagation (BP) or Feedback Alignment (FA)-like methods require separate error networks (red) for computing weight updates. 
            \textbf{b} Layer-wise GIM, with one layer per gradient-isolated module, does not transmit error signals across layers (i.e. modules) but requires (1) the transmission of information other than the network activity (red) and (2) a perfect replay of negative samples. 
            Thus, the resulting update computation needs a separate loss network and cannot be interpreted as a local learning rule.
            \textbf{c} Contrastive Local and Predictive Plasticity (CLAPP) calculates updates using locally and temporally available information: pre- and post-synaptic activity and predictive recurrent input onto the apical dendrite $\bm W^{\mathrm{pred}} \bm c^{t-\delta t}$.
            Global broadcasting factors $\gamma_t$ modulate plasticity depending on the presence or absence of a saccade. 
    }
    \label{fig:plasticity}
    \vspace{-0.5cm}
\end{figure}

\paragraph{Related work}
Many recent models of synaptic plasticity fit an apparently similar learning rule prototype \citep{lillicrap2016random,nokland2016direct,Roelfsema2018,Nokland2019,lillicrap2020backpropagation,Pozzi2020} if we interpret the top-down signals emerging from the BP algorithm as the dendritic signal.
However, top-down error signals in BP are not directly related to the activity $\bm c^t$ of the neurons in the main network during processing of sensory input.
Rather, they require a separate linear network mirroring the initial network and feeding back error signals (see \autoref{fig:plasticity} a and \cite{lillicrap2020backpropagation}), or involved time-multiplexing of feedforward and error signals in the main network \citep{lillicrap2020backpropagation, payeur2021burst}.
Our model is fundamentally different, because in our case, the dendritic signal onto neuron $j$ is strictly $(\bm W^{\mathrm{pred}} \bm c^t)_j$ which is a weighted sum of the main network activity and there is no need of a (linear) feedback network transmitting exact error values across many layers.

Moreover, we show in simulations in \autoref{sec:empirical_results}, that the dendritic signal does not have to come from a layer above but that the prediction fed to layer $l$ may come from the same layer. 
This shows that our learning rule works even in the complete absence of downward signaling from $l+1$ to $l$.
This last point is a significant difference to other methods that also calculate updates using only activities of the main network, but require tuned top-down connections to propagate signals downwards in the network hierarchy \citep{Kunin2020}, such as methods in the difference target propagation family \citep{lee2015difference, bartunov2018assessing, golkar2020biologically}, contrastive divergence \citep{hinton2002training} and equilibrium propagation \citep{scellier2017equilibrium}. Furthermore, the latter two require convergence to an equilibrium state for each input \citep{laborieux2021scaling}.
Our model does not require this convergence because it uses the recurrent dendritic signal $(\bm W^{\mathrm{pred}} \bm c^t)_j$ only for synaptic plasticity and not for inference.

Most previous learning rules which include global modulating factors interpret it as a reward prediction error \citep{schultz1997neural,gerstner2018eligibility,Pozzi2020}. 
In this paper, we address self-supervised learning and view global modulating factors as broadcasting signals, modeling the self-awareness that something has changed in the stimulus (e.g. because of a saccade).
Hence, the main function of the broadcast factor in our model is to identify contrastive inputs, which avoids a common pitfall for self-supervised learning models: `trivial' or `collapsed' solutions, where the model produces a constant output, which is easily predictable, but useless for downstream tasks.
In vision, we use a broadcast factor to model the strong, saccade-specific activity patterns identified throughout the visual pathway \citep{Kowler1995, leopold1998, ross2001changes, mcfarland2015}.
In other sensory pathways, like audition, this broadcast factor may model attention signals arising when changing focus on a new input source \citep{Fritz2007}, cross-modal input indicating a change in head or gaze direction, or signal/speaker-identity inferred from blind source separation, which can be done on low-level representation with biologically plausible learning rules \citep{hyvarinen1997fast, ziehe1998tdsep, molgedey1994separation}.
Our learning rule further requires this global factor to predict the absence or presence of a gaze change, hence conveying a \emph{change prediction error} rather than classical reward prediction error. 
Here, we do not model the precise circuitry computing this factor in the brain, however, we speculate that a population of neurons could express such a scalar factor e.g. through burst-driven multiplexing of activity, see \citet{payeur2021burst} and \autoref{app:extra}.


Our theory takes inspiration from the substantial progress seen in unsupervised machine learning in recent years and specifically from contrastive predictive coding (CPC) \citep{Oord2018}.
CPC trains a network (called {\it encoder}) to make {\it predictions} of its own responses to future inputs, while keeping this prediction as different as possible to its responses to {\it fake} inputs ({\it contrasting}). 
A key feature of CPC is that predicting and contrasting happens in latent space, i.e. on the output representation of the encoder network.
This avoids modeling a generative model for perfect reconstruction of the input and all its details (e.g. green, spiky). Instead the model is forced to focus on extracting high-level information (e.g. cactus).
In our notation, CPC evaluates a prediction $\bm W^{\mathrm{pred}} \bm c^{t}$ such that a score function $u_t^{\tau} = {\bm z^{\tau}}^\top  \bm W^{\mathrm{pred}} \bm c^{t}$ becomes larger for the true future $\tau=t+\delta t$ (referred to as positive sample) than for any other vector $\bm z^{t'}$ taken at arbitrary time points $t'$ elsewhere in the entire training set (referred to as negative samples in CPC).
This means, that the prediction should align with the future activity $\bm z^{t+\delta t}$ but not with the negative samples. \citet{Oord2018} formalizes this as a softmax cross-entropy classification, which leads to the traditional CPC loss:  
\begin{equation}
    \mathcal{L}_{\mathrm{CPC}}^t 
    = - \log \frac{\exp u_t^{t+\delta t}}{\sum_{\tau \in \mathcal{T}} \exp u_t^{\tau}}~, 
    \label{eq:cpc}
\end{equation}
where $\mathcal{T} = \left\{ t^{t+\delta t}, t'_1\dots t'_N\right\}$ comprises the positive sample and $N$ negative samples. 
The learned model parameters are the elements of the matrix $\bm W^{\mathrm{pred}}$, as well as the weights of the encoder network.
The loss function $\mathcal{L}_{\mathrm{CPC}}^t$ is then minimized by stochastic gradient descent on these parameters using BP.
Amongst numerous recent variants of contrastive learning \citep{He2019, Chen2020, Xiong2020}, 
we focus here on CPC \citep{Oord2018}, for which a more local variant, Greedy InfoMax, was recently proposed by \citet{Lowe2019}.

Greedy InfoMax (GIM) \citep{Lowe2019} is a variant of CPC which makes a step towards local, BP-free learning: the main idea is to split the encoder network into a few gradient-isolated modules to avoid back-propagation between these modules. 
As the authors mention in their conclusion, ``\emph{the biological plausibility of GIM is limited by the use of negative samples and within-module back-propagation}''.
This within-module back-propagation 
still requires a separate feedback network to propagate prediction errors (\autoref{fig:plasticity} a), 
but can be avoided in the most extreme version of GIM, where each gradient-isolated module contains a single layer ({\it layer-wise GIM}).
However, the gradients of layer-wise GIM, derived from \autoref{eq:cpc}, still cannot be interpreted as synaptic plasticity rules because the gradient computation requires
(1) the transmission of information other than the network activity (see \autoref{fig:plasticity} b), and 
(2) perfect memory to replay the negative samples $\bm z^{t'}$, as mentioned in the above quote (see \autoref{appendix:gradient} for details).
Overall it is not clear how this weight update of layer-wise GIM could be implemented with realistic neuronal circuits.
Our CLAPP rule solves the above mentioned implausibilities and allows a truly local implementation in space and time.

\section{Derivation of the CLAPP rule: contrastive, local and predictive plasticity}\label{sec:clapp_model}

We now suggest a simpler contrastive learning algorithm which solves the issues encountered with layer-wise GIM and for which a gradient descent update is naturally compatible with the learning rule prototype from \autoref{eq:rule_prototype}.
The most essential difference compared to CPC or GIM is, that we do not require the network to simultaneously access the true future activity $\bm z^{t+\delta t}$ and recall (or imagine) the network activity $\bm z^{t'}$ seen at some other time.
Rather, we consider the naturalistic time-flow illustrated in \autoref{fig:fig0} a, where an agent fixates on a moving animal for a while and then changes gaze spontaneously.
In this way, the prediction $\bm W^{\mathrm{pred}} \bm c^t$ is expected to be meaningful during fixation, but inappropriate right after a saccade. 
In our simulations, we model this by feeding the network with subsequent frames from the same sample (e.g. different views of an eagle), and then abruptly changing to frames from another sample (e.g. different views of an elephant).

We note that the future activity $\bm z^{t+\delta t}$ and the context $\bm c^t $ are \emph{always} taken from the main feedforward encoder network.
We focus on the case where the context stems from the same layer as the future activity ($\bm c^{t,l} = \bm z^{t,l}$), however, the model allows for the more general case, where the context stems from another layer (e.g. the layer above $\bm c^{t,l} = \bm z^{t,l+1}$).

\paragraph{Derivation of the CLAPP rule from a self-supervised learning principle}\label{sec:gradient_analysis}

Rather than using a global loss function for multi-class classification to separate the true future from multiple negative samples, as in CPC, we consider here a binary classification problem at every layer $l$: 
we interpret the score function $u_t^{t+\delta t, l} = {\bm z^{t+\delta t, l}}^\top \bm W^{\mathrm{pred},l} \bm c^{t, l}$ as the layer's `guess' whether the agent performed a fixation or a saccade. 
In \autoref{app:extra}, we discuss how $u_t^{t+\delta t, l}$ could be (approximately) computed in real neuronal circuits.
In short, every neuron $i$ has access to its `own' dendritic prediction $\hat{z}_i^{t,l} = \sum_j W^{\mathrm{pred},l}_{ij} c^{t,l}_j$ of somatic activity \citep{urbanczik2014learning}, and the product $z_i^{t+\delta t,l} \, \hat{z}_i^{t,l}$ can be seen as a coincidence detector of dendritic and somatic activity, communicated by specific burst signals \citep{Larkum1999}.
These burst signals allow time-multiplexed communication \citep{payeur2021burst} of the products  $z_i^{t+\delta t,l} \, \hat{z}_i^{t,l}$ of many neurons, which can then be summed by an interneuron representing $u_t^{t+\delta t, l}$.

As mentioned in \autoref{sec:related_work}, information about the presence or absence of a saccade between two time points is available in the visual processing stream and is modeled here by the variable $y^t=-1$ and $y^t=+1$, respectively. We interpret $y^t$ as the label of a binary classification problem, characterized by the Hinge loss, and define the CLAPP loss at layer $l$ as: 

\begin{eqnarray}
\mathcal{L}_{CLAPP}^{t, l} = \text{max}\left(0, 1 - y^t \cdot u_t^{t+\delta t, l} \right) ~~\text{with}~~~\left\{\begin{array}{ll}
       y^t=+1 & \mbox{for fixation} \\
       y^t=-1 & \mbox{for saccade}
    \end{array}
    \right.
\label{eq:CLAPP_loss}
\end{eqnarray}

We now derive the gradients of \autoref{eq:CLAPP_loss} with respect to the feedforward weights 
and show that gradient descent on this loss function is compatible with the learning rule prototype suggested in \autoref{eq:rule_prototype}.
Note that CLAPP optimises \autoref{eq:CLAPP_loss} for each layer $l$ independently, without any gradient flow between layers.
That being said, the following derivation is the same for every layer $l$, which is why we omit the layer index $l$ from here on.

Since we chose to formalize the binary classification with a Hinge loss, the gradient vanishes when the classification is already correct: high score $u_t^{t+\delta t} > 1$ during fixation ($y^t=+1$), or a low score $u_t^{t+\delta t} < -1$ after a saccade ($y^t=-1$). Otherwise, it is $-\nabla u^{t+\delta t}_t$ during a fixation or $\nabla u^{t+\delta t}_t$ after a saccade.
In the `predicted layer' $\bm z$, i.e. the target of the prediction, let $W_{ji}$ denote the feedforward weight from neuron $i$ in the previous layer (with activity $x^{t}_{i}$) to neuron $j$, with summed input $a_j^t$ and activity $z_j^t$.
Similarly, in the `predicting layer' $\bm c$, i.e. the source of the prediction, let $W^{{\bm c}}_{kl}$ denote the feedforward weight between the neuron $l$ in the previous layer (with activity $x^{\bm c,t}_l$) and neuron $k$, with summed input $a_k^{\bm c, t}$ and activity $c^t_k$.
Therefore, $\cdot^{\bm c}$ as an upper index refers to the context layer, whereas $\bm c$ as a full-size letter refers to the respective neuronal activity.
We then find the gradients with respect to these weights as:

\begin{eqnarray}
    \frac{\partial \LL_{CLAPP}^t }{\partial  W_{ji}} & = &
    \pm (\bm W^\mathrm{pred} \bm c^t)_j  ~ \rho'(a_j^{t+\delta t}) ~ x_i^{t+\delta t}
    \label{eq:grad-clapp-1}
    \\
    \frac{\partial \LL_{CLAPP}^t }{\partial  W_{km}^{\bm{c}}} &= & \pm ( {\bm W^\mathrm{pred}}^{\top} \bm z^{t+\delta t})_k
    ~ \rho '(a_k^{\bm c,t})
    ~ x_m^{\bm c,t}~,
    \label{eq:grad-clapp-2}
\end{eqnarray}
where the sign is negative during fixation and positive after a saccade.
To change these equations into online weight updates, we consider the gradient descent update delayed by $\delta t$, such that
$\Delta W_{ji}^{t} = - \eta
\frac{\partial \LL_{CLAPP}^{t-\delta t}}{\partial  W_{ji}}$,
where $\eta$ is the learning rate.
Let us define a modulating factor $\gamma_t = y^t \cdot H^t$, where $y^t = \pm 1$ is a network-wide broadcast signal (self-awareness) indicating a saccade ($-1$) or a fixation ($+1$) and $H^t \in \{0,\eta\}$ is a layer-wide broadcast signal indicating whether the saccade or fixation was correctly classified as such.
In this way, \autoref{eq:grad-clapp-1} becomes a weight update which follows strictly the ideal learning rule prototype from \autoref{eq:rule_prototype}:
\begin{equation}
\Delta W_{ji}^{t} = \underbrace{~\gamma_t~}_{\text{broadcast factors}}
\cdot
\underbrace{(\bm W^\mathrm{pred} \bm c^{t-\delta t})_j}_{\text{dendritic prediction}}
\cdot
\underbrace{\rho'(a_j^{t}) x_i^{t}}_{\text{local activity}}~.
\label{eq:dw_z_CLAPP}
\end{equation}
For the updates of the connections onto the neuron $c_k^t$, which emits the prediction rather than receiving it, our theory in \autoref{eq:grad-clapp-2} requires the opposite temporal order and the transmission of the information in the opposite direction: from $\bm z^t$ back to $\bm c^t$.
Since connections in the brain are unidirectional \citep{lillicrap2016random}, we introduce another matrix $\bm W^{\mathrm{retro}}$ which replaces ${\bm W^{\mathrm{pred}}}^\top$ in the final weight update. 
Given the inverse temporal order, we interpret $\bm W^{\mathrm{retro}} \bm z^t$ as a retrodiction rather than a prediction. 
In \autoref{app:extra}, we show that using $\bm W^{\mathrm{retro}}$ minimises a loss function of the same form as \autoref{eq:CLAPP_loss}, and empirically performs as well as using ${\bm W^{\mathrm{pred}}}^\top$.
The resulting weight update satisfies the learning rule prototype from \autoref{eq:rule_prototype}, as it can be written:
\begin{equation}
\Delta W_{km}^{\bm c,t} = \underbrace{~\gamma_t~}_{\text{broadcast factors}} \cdot
\underbrace{(\bm W^\mathrm{retro} \bm z^{t})_k}_{\text{dendritic retrodiction}}
\cdot
\underbrace{\rho'(a_k^{\bm c, t-\delta t}) x_m^{\bm c, t-\delta t}}_{\text{local activity}}~.
\label{eq:dw_c_CLAPP}
\end{equation}
In the (standard) case, where context and predicted activity are from the same layer ($\bm c^{t,l} = \bm z^{t,l}$), $\bm W$ and $\bm W^{\bm c}$ are the same weights and the updates \autoref{eq:dw_z_CLAPP} and \autoref{eq:dw_c_CLAPP} are added up linearly.

The prediction and retrodiction weights, $\bm W^{\mathrm{pred}}$ and $\bm W^{\mathrm{retro}}$, respectively, are also plastic. 
By deriving the gradients of $\LL_{CLAPP}^t$ with respect to $\bm W^{\mathrm{pred}}$, we find an even simpler Hebbian learning rule for these weights:
\begin{equation}
    \Delta W^{\mathrm{pred}}_{jk}
    =
    \Delta W^{\mathrm{retro}}_{kj}
    = 
    \underbrace{\gamma_{t}}_{\text{broadcast factors}} \cdot ~
    \underbrace{\textcolor{black}{z_j^{t}} ~ \cdot ~ \textcolor{black}{c_k^{t-\delta t}}}_{\text{\textcolor{black}{pre~and~post}}} ~,
    \label{eq:dw_pred_CLAPP}
\end{equation}

where neuron $k$ in the predicting layer $\bm c$ is pre-synaptic (post-synaptic) and neuron $j$ in the predicted layer $\bm z$ is post-synaptic (pre-synaptic) for the prediction weights $W^{\mathrm{pred}}_{jk}$ (retrodiction weights $W^{\mathrm{retro}}_{kj}$).
Note that the update rules for $W^{\mathrm{pred}}_{jk}$ and $W^{\mathrm{retro}}_{kj}$ are reciprocal, a method that leads to mirrored connections, given small enough initialisation \citep{burbank2015mirrored, amit2019deep, Pozzi2020}.

We emphasize that all information needed to calculate the above CLAPP updates (Equations \ref{eq:dw_z_CLAPP} -- \ref{eq:dw_pred_CLAPP}) is spatially and temporally available, either as neuronal activity at time $t$, or as traces of recent activity ($t-\delta t$) \citep{gerstner2018eligibility}. 
In order to implement \autoref{eq:dw_z_CLAPP}, the dendritic prediction has to be retained during $\delta t$. However, we argue that dendritic activity can outlast (50-100 ms) somatic neuronal activity (2-10 ms) \citep{Major2013}, which makes predictive input from several time steps in the past ($t-\delta t$) available at time $t$.

\paragraph{Generalizations}

While the above derivation considers fully-connected feedforward networks, we apply analogous learning rules to convolutional neural networks (CNN) and recurrent neural networks (RNN).
Analyzing the biological plausibility of the standard spatial weight sharing and spatial MeanPooling operations in CNNs is beyond the scope of the current work.
Furthermore, we discuss in \autoref{app:extra}, that MaxPooling can be interpreted as a simple model of lateral inhibition and that gradient flow through such layers is compatibility with the learning rule prototype in \autoref{eq:rule_prototype}.

To obtain local learning rules even for RNNs, we combine CLAPP with the e-prop theory \citep{bellecsolution}, which provides a biologically plausible alternative to BP through time: gradients can be propagated forward in time through the intrinsic neural dynamics of a neuron using eligibility traces. The propagation of gradients across recurrently connected units is forbidden and disabled. This yields biologically plausible updates in GRU units, as explained in \autoref{app:extra}.

\begin{figure}[t]
    \centering
    \includegraphics[width=\textwidth]{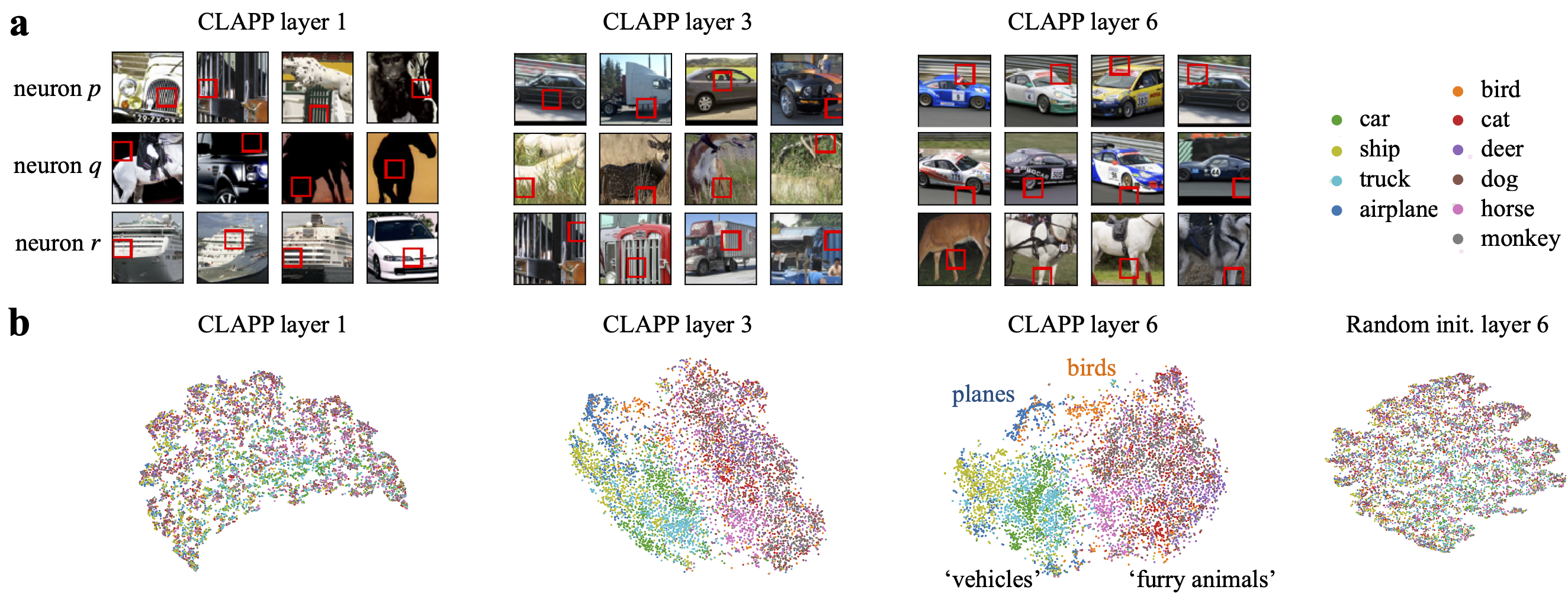}
    \caption{Hierarchical representations learned by CLAPP. \textbf{a} Red boxes in STL-10 images indicate patches that best activate a specific neuron (rows) in a network trained with CLAPP. Layer 1 extracts simple features like gratings or uniform patches, higher layers extract richer features like parts of objects. \textbf{b} 2-dimensional t-SNE projection of neuronal activities at different layers unveils increasing representational structure in higher layers (every dot represents one input image). Note that CLAPP has not seen any class labels during training.}
    \vspace{-0.5cm}
    \label{fig:vis}
\end{figure}

\section{Empirical results}\label{sec:empirical_results}

\paragraph{Building hierarchical representations}
We first demonstrate numerically, that CLAPP yields deep hierarchical representations, despite using a local plasticity rule compatible with \autoref{eq:rule_prototype}.
We report here the results for $\bm c^{t,l} = \bm z^{t,l}$, i.e. the dendritic prediction in \autoref{eq:rule_prototype} is generated from lateral connections and the representations in the same layer.
We note, however, that we obtained qualitatively similar results with $\bm c^{t,l} = \bm z^{t,l+1}$ (i.e. the dendritic prediction is generated from one layer above), suggesting that top-down signaling is neither necessary for, nor incompatible with, our algorithm (also see \autoref{app:extra}).


We first consider the STL-10 image dataset \citep{Coates2011}. 
To simulate a time dimension in these static images, we follow \cite{Henaff2019} and \cite{Lowe2019}: each image is split into $16 \times 16$ patches and the patches are viewed one after the other in a vertical order (one time step is one patch).
Other hyper-parameters and data-augmentation are taken from \citet{Lowe2019}, see \autoref{app:details}.
We then train a 6-layer VGG-like \citep{Simonyan2015} encoder (VGG-6) using the CLAPP rule (Equations \ref{eq:dw_z_CLAPP} -- \ref{eq:dw_pred_CLAPP}).
Training is performed on the unlabelled part of the STL-10 dataset for 300 epochs. We use 4 GPUs (\texttt{NVIDIA Tesla V100-SXM2 32 GB}) for data-parallel training, resulting in a simulation time of around 4 days per run. 

In order to study how neuronal selectivity changes over layers, we select neurons randomly and show image patches which best activate these neurons the most (rows in \autoref{fig:vis} a).
As expected for a visual hierarchy, first-layer neurons (first column in \autoref{fig:vis} a) are selective to horizontal or vertical gratings, or homogeneous colors.
In the third layer of the network (second column), neurons start to be selective to more semantic features like grass, or parts of vehicles. Neurons in the last layer (third column) are selective to specific object parts (e.g. a wheel touching the road).
The same analysis for a random, untrained encoder does not reveal a clear hierarchy across layers, see \autoref{app:extra}.

To get a qualitative idea of the learned representation manifold, we use the non-linear dimension reduction technique t-SNE \citep{van2008visualizing} to visualise the encodings of the (labeled) STL-10 test set in \autoref{fig:vis} b.
We see that the representation in the first layer is mostly unrelated to the underlying class.
In the third and sixth layers' representation, a coherent clustering emerges, yielding an almost perfect separation between furry animals and vehicles.
This clustered representation is remarkable since the network has never seen class labels, and was never instructed to separate classes, during CLAPP training
The representation of the same architecture, but without training (Random init.), shows that a convolutional architecture alone does not yield semantic features.

\begin{SCfigure}
    \centering
	\includegraphics[width=0.5\textwidth]{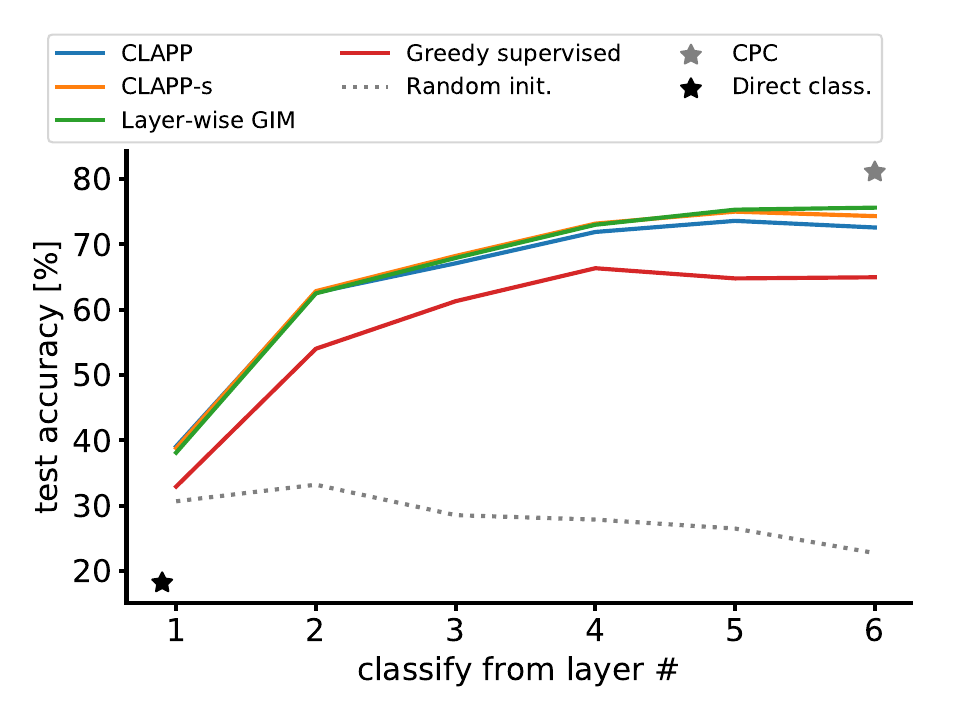}
	\captionof{figure}{
            CLAPP stacks well: representations after stacking up to 5 layers increase performance of a linear classifier on STL-10, despite the local learning rule (blue and orange lines), while performance decreases for convolutional network with weights fixed at random initialisation (dotted).
            Greedy supervised training (see \autoref{app:details}) also stacks, but already saturates at layer 4 and shows overall lower performance.
            Direct linear classification on image pixels (black star) and CPC performance after 6 layers (gray star) serve as upper and lower performance bounds, respectively. 
    }
    \label{fig:results}
    \vspace{-0.3cm}
\end{SCfigure}

To produce a more quantitative measurement of the quality of learned representations, we follow the methodology of \cite{Oord2018} and \cite{Lowe2019}: we freeze the trained encoder weights and train a linear classifier to recognize the class labels from each individual layer (\autoref{fig:results}).
As expected for a deep representation, the classification accuracy increases monotonically with the layer number and only saturates at layers $5$ and $6$.
The accuracies obtained with layer-wise GIM are almost indistinguishable from those obtained with CLAPP.
It is only at the last two layers, that layer-wise GIM performs slightly better than CLAPP; yet GIM has multiple biologically implausible features that are removed by CLAPP.
As a further benchmark, we also plot the accuracies obtained with an encoder trained with greedy supervised training. 
This method trains each layer independently using a supervised classifier at each layer, without BP between layers, which results in an almost local update (see \citet{Lowe2019} and \autoref{app:details}). 
We find that accuracy is overall lower and saturates already at layer $4$. On this dataset, with many more unlabelled than labelled images, greedy supervised accuracy is almost $10 \%$ below the accuracy obtained with CLAPP.
Again, we see that a convolutional architecture alone does not yield hierarchical representations, as performance decreases at higher layers for a fixed random encoder.


\paragraph{Comparing CPC and CLAPP}
Since CLAPP can be seen as a simplification of CPC (or GIM) 
we study four algorithmic differences between CPC and CLAPP individually. They are:
(1) Gradients in CLAPP (layer-wise GIM) cannot flow from a layer to the next one, as opposed to BP in CPC, 
(2) CLAPP performs a binary comparison (fixation vs. saccade) with the Hinge loss, whereas CPC does multi-class classification with the cross entropy loss,
(3) CLAPP processes a single input at a time, whereas CPC uses many positive and negative samples synchronously, and 
(4) we introduced $\bm W^{\mathrm{retro}}$ to avoid the weight transport problem in $\bm W^{\mathrm{pred}}$.

We first study 
features (1) and (2) but relax constraints (3) and (4).
That means, in this paragraph, we allow a fixation and $N=16$ synchronous saccades and set $\bm W^{\mathrm{retro}} = \bm W^{\mathrm{pred},\top}$.
We refer to \emph{Hinge Loss CPC} as the algorithm minimizing the CLAPP loss (\autoref{eq:CLAPP_loss}) but using end-to-end BP.
\emph{CLAPP-s} (for \emph{synchronous}) applies the Hinge Loss to every layer, but with gradients blocked between layer.
We find that the difference between the CPC loss and Hinge Loss CPC is less than $1\%$, see \autoref{tab:comp}. 
In contrast, additional blocking of gradients between layers causes a performance drop of almost $5\%$ for both loss functions.
We investigate how gradient blocking influences performance with a series of simulations, splitting the $6$ layers of the network into two or three gradient isolated modules, exploring the transition from Hinge Loss CPC to CLAPP-s. 
Performance drops monotonously but not catastrophically, as the number of gradient blocks increases (\autoref{tab:comp}).

CLAPP's temporal locality allows the interpretation that an agent alternates between fixations and saccades, rather than perfect recall and synchronous processing of negative samples, as required by CPC.
To study the effect of temporal locality, we apply features (2) and (3) and relax the constraints (1) and (4).
The algorithm combining temporal locality and the CLAPP loss function is referred to as \emph{time-local Hinge Loss CPC}. 
We find that the temporal locality constraint decreases accuracy by $1.2\%$ compared to Hinge Loss CPC.
The last feature introduced for biological plausibility is using the matrix $\bm W^\mathrm{retro}$ and we observe almost no difference in classification accuracy with this alternative (the accuracy decreases by $0.1\%$).
Conversely, omitting the update in \autoref{eq:dw_c_CLAPP} entirely, i.e. setting the retrodiction $\bm W^\mathrm{retro} = 0$, compromises accuracy by $2\%$ compared to vanilla Hinge Loss CPC.

When combining all features (1) to (4), we find that the fully local CLAPP learning rule leads to an accuracy of $73.6\%$ at layer $5$. We conclude from the analysis above, that the feature with the biggest impact on performance is (1): blocking the gradients between each layer.
However, despite the performance drop caused by blocking the gradients, CLAPP still stacks well and leverages the depth of the network (\autoref{fig:results}).
All other features (2) - (4), introduced to derive a weight update compatible with our prototype (\autoref{eq:rule_prototype}), only caused a minor performance loss.

\begin{table}[t]
    \centering
	\caption{
		    CLAPP performs best among methods that are local in space and time.
		    Linear classification test accuracy [\%] on STL-10, phone classification on LibriSpeech, and video human action recognition on UCF-101 using features from the encoder trained with different methods.
		    On STL-10, performance 
		    degrades gracefully with the number of gradient-isolated modules in the VGG-6 encoder (at fixed number of encoder layers).
		    Greedy supervised training uses BP in auxiliary classifier networks (`almost' local in space). 
		    For LibriSpeech, BP through time is used (can be avoided, see \autoref{app:extra}).
		    Values with * are taken from \citet{Lowe2019}. For simulation details, see \autoref{app:details}. 
	}
	\vspace{0.2cm}
	\begin{tabular}{lccccc} 
         \hline
         \multirow{2}{*}{Method} & \multicolumn{2}{c}{local in \dots}  & \multirow{2}{*}{STL-10} & \multirow{2}{*}{LibriSpeech} & \multirow{2}{*}{UCF-101} \\
         & space? & time? &&&\\\hline\hline
         Chance performance &&& 10.0 & 2.4 & 0.99 \\\hline
         Random init. & \textcolor{green}{\ding{51}} & \textcolor{green}{\ding{51}} & 21.8 & 27.7* & 30.5\\ 
         MFCC & \textcolor{green}{\ding{51}} & \textcolor{green}{\ding{51}} & - & 39.7* & -\\\hline
         Greedy supervised & (\textcolor{green}{\ding{51}}) & \textcolor{green}{\ding{51}} & 66.3 & 73.4* & -\\
         Supervised & \textcolor{red}{\ding{55}} & \textcolor{green}{\ding{51}} & 73.2 & 77.7* & 51.5\\\hline
         CPC & \textcolor{red}{\ding{55}} & \textcolor{red}{\ding{55}} & 81.1 & 64.3 & 35.7\\
         Layer-wise GIM & \textcolor{red}{\ding{55}} & \textcolor{red}{\ding{55}} & 75.6 & 63.9 & 41.2\\\hline\hline
         Hinge Loss CPC (ours)\hspace{-0.9cm} & \textcolor{red}{\ding{55}} & \textcolor{red}{\ding{55}} & 80.3 & 62.8 & 36.1\\
         CLAPP-s (2 modules of 3 layers) & \textcolor{red}{\ding{55}} & \textcolor{red}{\ding{55}} & 77.6 & - & -\\
         CLAPP-s (3 modules of 2 layers) & \textcolor{red}{\ding{55}} & \textcolor{red}{\ding{55}} & 77.4 & - & -\\
         CLAPP-s (ours) & \textcolor{green}{\ding{51}} & \textcolor{red}{\ding{55}} & 75.0 & 61.7 & 41.6\\\hline
         time-local Hinge Loss CPC (ours) & \textcolor{red}{\ding{55}} & \textcolor{green}{\ding{51}} & 79.1 & -& - \\
         CLAPP (ours) & \textcolor{green}{\ding{51}} & \textcolor{green}{\ding{51}} & 73.6 & - & - \\\hline
    \end{tabular}
    \label{tab:comp}
    \vspace{-0.4cm}
\end{table}

\paragraph{Applying CLAPP to speech and video} We now demonstrate that CLAPP is applicable to other modalities like the LibriSpeech dataset of spoken speech \citep{Panayotov2015} and the UCF-101 dataset containing short videos of human actions \citep{soomro2012ucf101}.
When applying CLAPP to auditory signals, we do not explicitly model the contrasting mechanism (saccades in the vision task; see discussion in \autoref{sec:related_work} for the auditory pathway) and hence consider the application of CLAPP as benchmark application, rather than a neuroscientifically exact study.
To increase computational efficiency, we study CLAPP-s on speech and video.
Based on the image experiments, we expect similar results for CLAPP, given enough time to converge.
We use the same performance criteria as for the image experiments and summarize our results in \autoref{tab:comp}, for details see \autoref{app:details}.

For the audio example, we use the same architecture as \citet{Oord2018} and \cite{Lowe2019}: multiple temporal 1d-convolution layers and one recurrent GRU layer on top. 
As in the feedforward case, CLAPP still optimises the objective of \autoref{eq:CLAPP_loss}.
For the 1d-convolution layers, the context $\bm c^t$ is computed as for the image task, for the last layer, $\bm c^t$ is the output of the recurrent GRU layer.
We compare the performance of the algorithms on phoneme classification (41 classes) using labels provided by \cite{Oord2018}. 
In this setting, layer-wise training lowers performance by only $0.4\%$ for layer-wise GIM, and by $1.1\%$ for CLAPP-s. 
Implemented as such, 
CLAPP-s still relies on BP through time (BPTT) to train the GRU layer. 
Using CLAPP-s with biologically plausible e-prop \citep{bellecsolution}, instead of non-local BPTT, reduces performance by only 3.1 \%, whereas omitting the GRU layer compromises performance by 9.3 \%, see \autoref{app:extra}.

Applying CLAPP to videos is especially interesting because their temporal sequence of  images perfectly fits the scenario of \autoref{fig:fig0} a.
In this setting, we take inspiration from \citet{Han2019}, and use a VGG-like stack of 2D and 3D convolutions to process video frames over time.
On this task (101 classes), we found layer-wise GIM and CLAPP-s to achieve higher downstream classification accuracy than their end-to-end counterparts CPC and Hinge Loss CPC (see \autoref{tab:comp}), in line with the findings on STL-10 in \citet{Lowe2019}.
On the other hand, we found that CLAPP-s requires more negative samples (i.e. more simultaneous comparisons of positive and negative samples) on videos than on STL-10 and LibriSpeech. 
Under the constraint of temporal locality in fully local CLAPP, this leads to prohibitively long convergence times in the current setup.
However, since CLAPP linearly combines updates stemming from multiple negative and positive samples, we eventually expect the same final performance, if we run the online CLAPP algorithm for a sufficiently long time.

\section{Discussion}\label{sec:discussion}

We introduced CLAPP, a self-supervised and biologically plausible learning rule that yields deep hierarchical representations in neural networks.
CLAPP integrates neuroscientific evidence on the dendritic morphology of neurons and takes the temporal structure of natural data into account. 
Algorithmically, CLAPP minimises a layer-wise contrastive predictive loss function
and stacks well on different task domains like images, speech and video -- despite the locality in space and time.

While the performance loss due to layer-wise training is a limitation of the current model, the stacking property is preserved and preliminary results suggest improved versions that stack even better (e.g. using adaptive encoding patch sizes).
Note that CLAPP models self-supervised learning of cortical hierarchies and does \emph{not} provide a general credit assignment method, such as BP. 
However, the representation learned with CLAPP could serve as an initialisation for transfer learning, where the encoder is fine-tuned later with standard BP.
Alternatively, fine-tuning could even start already during CLAPP training.
CLAPP in its current form is data- and compute-intensive, however, it runs on unlabelled data with quasi infinite supply, and is eligible for neuromorphic hardware, which could decrease energy consumption dramatically \citep{wunderlich2019demonstrating}.

Classical predictive coding models alter neural activity at inference time, e.g. by cancelling predicted future activity \citep{Rao1999, Keller2018}.
Here, we suggest a different, perhaps complementary, role of predictive coding in synaptic plasticity, where dendritic activity predicts future neural activity, but directly enters the learning rule \citep{Kording2001, urbanczik2014learning}.
CLAPP currently does not model certain features of biological neurons, e.g. spiking activity or long range feedback, and requires neurons to transmit signals with precise value and timing.
We plan to address these topics in future work.

\begin{ack}
This research was supported by the Swiss National Science Foundation (no. 200020\_184615) and the Intel Neuromorphic Research Lab. Many thanks to Sindy L\"owe, Julie Grollier, Maxence Ernoult, Franz Scherr, Johanni Brea and Martin Barry for helpful discussions. Special thanks to Sindy L\"owe for publishing the GIM code.
\end{ack}

\bibliographystyle{abbrvnat} 
\bibliography{references}

\newcommand{\toptitlebar}{
  \hrule height 4pt
  \vskip 0.25in
  \vskip -\parskip%
}
\newcommand{\bottomtitlebar}{
  \vskip 0.29in
  \vskip -\parskip
  \hrule height 1pt
  \vskip 0.09in%
}

\newpage
\setcounter{page}{1}
\appendix
\begin{center}
{\Large \textbf{Appendices of:\\}}
\vspace{0.5cm}

\toptitlebar
{\LARGE \textbf{\newtitle}}
\bottomtitlebar
\vspace{0.5cm}
\end{center}

\paragraph{Notation in appendices} 
In all appendices, and in line with \citep{Oord2018,Lowe2019}, the context vector, from which the prediction is performed, is denoted $\bm c^t$ and the feature vector being predicted is denoted $\bm z^{t+\delta t}$ (or $\bm z^{t'}$ for negative samples).
In general, the loss function of CPC and CLAPP are therefore defined with the score functions $u_t^{\tau} = \bm z^{\tau} {}^\top \bm W^{\mathrm{pred}} \bm c^{t}$.

Throughout the vision experiments and when training the temporal convolutions of the audio processing network, it happens that $\bm c$ and $\bm z$ denote the same layer (see \autoref{app:details} for details). However, when processing audio, the highest loss uses the last layer as the context layer $\bm c$ and the one before last for $\bm z$.

To cover the most general case, we introduce different notations for the parameters and the variables of the context layer $\bm c$ and the feature layer $\bm z$.
For simplicity our analysis considers standard, fully-connected networks -- even if the reasoning generalises easily to other architectures. Hence, with a non-linearity $\rho$, the feature layer produces the activity $\bm z^{t}=\rho(\bm a^{\bm z,t })$ with $\bm a^{\bm z,t } = \bm W^{\bm z} \bm x^{\bm z,t} + \bm b^{\bm z}$ where $\bm x^{\bm z,t}, \bm W^{\bm z}$ and $\bm b^{\bm z}$ are the input vector (at time $t$), weight matrix and bias respectively (the layer index $l$ is omitted for simplicity). The notation naturally extends to the context layer $\bm c$ and we use $\bm x^{\bm c,t}, \bm W^{\bm c}$ and $\bm b^{\bm c}$ to denote its input and its parameters. Note that when the context and feature layer are the same layer $\bm z = \bm c$, the two parameters $\bm W^{c}$ and $\bm W^{z}$ are actually only one single parameter $\bm W$ and the weight update is given by $\Delta \bm W = \Delta \bm W^{c} + \Delta \bm W^{z}$.

For the gradient computations in the appendices we assume that the gradient cannot propagate further than one layer. Hence, $\bm x^{z}$ and $\bm x^{\bm c}$ are always considered as constants with respect to all parameters, even though this is technically not true, for instance with $\bm c^{l} = \bm z^{l+1}$. In this case we would have $\bm z = \bm x^{\bm c}$ and thus $\nabla_{\bm {W}^{\bm z}} \bm x^{\bm c} \neq \bm 0$, but we use the convention $\nabla_{\bm {W}^{\bm z}} \bm x^{\bm c} = \bm 0$ to obtain local learning rules. Gradients are computed accordingly by stopping gradient propagation in all our experiments.

\begin{figure}[h]
    \centering
    \includegraphics[width=\textwidth]{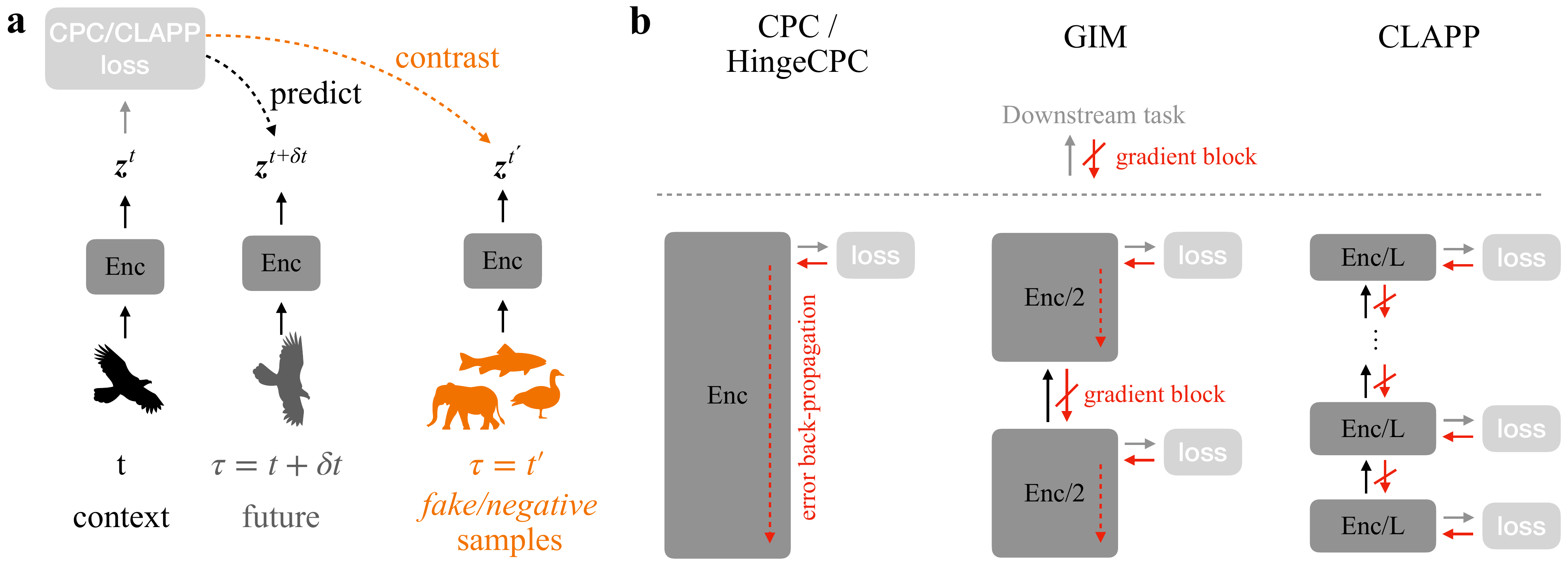}
    \caption{
            \textbf{a} In Contrastive Predictive Coding (CPC) and CLAPP(-s), an encoder network (Enc) produces a representation $\bm z^t$ at time $t$ (sometimes more generally called `context'). 
            Given  $\bm z^t$, the encoding of the future input $\bm z^{t+\delta t}$ should be {\it predicted} while keeping the prediction as different as possible from encoded {\it fake} or {\it negative} samples $\bm z^{t'}$ ({\it contrasting}).
            The loss function implementing this contrasting depends on the method: CPC uses cross-entropy classification, CLAPP uses a Hinge-loss.
            \textbf{b} CPC trains the encoder network end-to-end using gradient back-propagation (red arrows). In Greedy InfoMax (GIM), the encoder network is split into several, gradient-isolated modules and the loss (CPC or Hinge) is applied separately to each module. Gradient back-propagation still occurs within modules (red, dashed arrows) but is blocked between modules. In CLAPP, every module contains only a single trainable layer of the $L$-layer encoder. This avoids any back-propagation and makes CLAPP layer-local.
    }
    \vspace{-0.5cm}
    \label{fig:fig1}
\end{figure}

\section{Analysis of the original CPC gradient}\label{appendix:gradient}

Even after preventing gradients to flow from a layer to the next, we argue that parts of the gradient computation in CPC and GIM are hard to implement with the type of information processing that is possible in neural circuits. For this reason we analyse the actual gradients computed by layer-wise GIM. We further discuss the bio-plausibility of the resulting gradient computation in this section.

To derive the loss gradient we define the probability $\pi_{t}^{t*}$ that the sample $\bm z^{t*}$ is predicted as the true future given the context layer $\bm c^{t}$: $\pi_t^{t*} \eqdef \frac{1}{\mathcal{Z}} \exp u_t^{t*}$ with $\mathcal{Z} \eqdef \sum_{\tau \in \mathcal{T}} \exp u_t^{\tau}$. The set $\mathcal{T} = \left\{ t^{t+\delta t}, t'_1\dots t'_N\right\}$ comprises the positive and $N$ negative samples. We have in particular $\mathcal{L}_{CPC}^t = - \log \pi_t^{t + \delta t}$ and for any parameter $\theta$ the (negative) loss gradient is given by:


\begin{eqnarray}
\nabla_\theta \log \pi_t^{t+\delta t} & = &
\nabla_\theta u_t^{t+\delta t} -
\sum_{\tau \in \mathcal{T}} \pi_t^{\tau}~ \nabla_\theta u_t^{\tau}~.
\label{eq:3}
\end{eqnarray}

We consider only three types of parameters: the weights $\bm W^{c}$ onto the context vector $\bm c^t$, the weights $\bm W^{z}$ onto the feature vector $\bm z^{t*}$ and the weights $\bm W^{\mathrm{pred}}$ defining the scalar score $u_t^{t*} =  {\bm z^{t*}}^\top  \bm W^{\mathrm{pred}} \bm c^{t}$ (the biases are absorbed in the weight matrices for simplicity).

Let's first analyze the gradient with respect to $\bm W^{\mathrm{pred}}$.
Using the conventions that $k$ is the index of the context unit $c_k$ and $j$ is the index of the feature unit $z_j$, we have:

\begin{eqnarray}
\nabla_{W_{jk}^{\mathrm{pred}}} \log \pi_t^{t+\delta t} 
& = & c_k^t
\left( z_j^{t+\delta t} - \sum_{\tau \in \mathcal{T}} \pi_t^{\tau}~ z_j^{\tau} \right) \label{eq:grad_w_pred}
\end{eqnarray}

Viewing a gradient descent weight update of that parameter as a model of synaptic plasticity in the brain raises essential questions. If $z_j^{t+\delta t} - \sum_{\tau \in \mathcal{T}} \pi_t^{\tau} z_j^{\tau}$ was the activity of the unit $j$, it would boil down to a Hebbian learning rule, well supported experimentally, but the activity of unit $j$ is considered to be the vector element $z_j$ since it is transmitted to the layer above during inference. Hence, the unit $j$ would have to transmit two distinct quantities at the same time, which is unrealistic when modelling real neurons. On top of that, it is unclear how the term $\sum_{\tau \in \mathcal{T}} \pi_t^{\tau} z_i^{\tau}$ would be computed. 

We now compute the gradient with respect to $\bm W^{\bm c}$ and $\bm W^{\bm z}$.
The update of these parameters raises an extra complication because it involves the activity of more than two units. 
For the parameters of the layer $\bm z$ we denote $j$ a neuron in this layer, and $i$ a neuron from its input layer $\bm x$. Then the loss gradient is given by:
\begin{eqnarray}
\nabla_{W_{ji}^z} \log \pi_t^{t+\delta t} 
& = & (\bm W^{\mathrm{pred}} \bm c^t)_j \left( \rho'(a_j^{\bm z})^{t+\delta t} x^{\bm z,t+\delta t}_i - \sum_{\tau \in \mathcal{T}} \pi_t^{\tau}~ \rho'(a_j^{\bm z})^{\tau}  x^{\bm z, \tau}_i \right) ~ . \label{eq:grad_w_z}
\end{eqnarray}
Similarly, for the parameters of a neuron $\bm c_j^t$:
\begin{eqnarray}
\nabla_{W_{ji}^c} \log \pi^{t+\delta t} 
& = & \left( \bm W^\mathrm{pred,\top}\left( \bm z^{t+\delta t} - \sum_{\tau \in \mathcal{T}} \pi_t^{\tau}~ \bm z^{\tau} \right) \right)_j ~ \rho'(a_j^{\bm c})^{t}  x^{\bm c, t}_i ~ . \label{eq:grad_w_c}
\end{eqnarray}
These gradients raise the same essential problems as the computation of the gradients with respect to $\bm W^{pred}$ and even involve other complex computations.

\section{Simulation details}\label{app:details}

We use pytorch \citep{Paszke2017} for our implementation and base it on the code base of the GIM paper \citep{Lowe2019} \footnote{\url{https://github.com/loeweX/Greedy_InfoMax}}. 
Unless mentioned otherwise we adopt their setup, data sets, data handling and (hyper-)parameters.

\subsection{Vision experiments}

\paragraph{General procedure}

We use the STL-10 dataset, designed for unsupervised learning algorithms \citep{Coates2011}, which contains $100,000$ unlabeled color images of $96 \times 96$ pixels. Additionally, STL-10 contains a much smaller labeled training set with 10 classes and 500 training images per class and 800 labeled test images per class.
Since CPC-like methods rely on sequences of data we have to introduce an artificial `temporal' dimension in the case of vision data sets.
To simulate a time dimension in these static images we represent the motion of the visual scene by splitting the image into partially overlapping tiles. 
Then, vertical slices of patches define a temporal order, as in \cite{Henaff2019} and \cite{Lowe2019}: the patches are viewed one after the other in a vertical order (one time step is one patch). 
The hyper-parameters of this procedure and of any other image preprocessing and data augmentation steps are as in \citet{Lowe2019}.

This results in a time varying input stimulus which is fed into the encoder network and the weights of this network are updated using the CLAPP rule \autoref{eq:dw_z_CLAPP} and \autoref{eq:dw_c_CLAPP} (or reference algorithms, respectively).
CLAPP represents saccades towards a new object by changing the next input image to a different one at any time step with probability $0.5$.
Note that this practice reduces the number of training data by $50 \%$ compared to CLAPP-s, GIM and CPC, which are updated with positive and negative sample synchronously at every step.
Since this slows down convergence, we grant CLAPP double the amount of training epochs to yield a fair comparison ($1 \%$ improvement for CLAPP).
We leverage common practices from deep learning to accelerate the simulation: the weight changes are averaged and applied after going through a batch of 32 images so that the images can be processed in parallel. We accumulate the gradient updates and use the Adam optimiser with fixed learning rate 0.0002.

We then freeze the encoder network and train a linear downstream classifier on representations created by the encoder using held-out, labeled data from 10 different classes from the STL-10 dataset. 
The accuracy of that classification serves as a measure to evaluate the quality of the learned encoder representations.

\paragraph{Encoder architecture}
We use VGG-6, a custom 6-layer VGG-like \citep{Simonyan2015} encoder with 6 trainable layers (6 convolutional, 4 MaxPool, 0 fully-connected, see \autoref{tab:vgg6-arch}). The architecture choice was inspired by the condensed VGG-like architectures successfully applied in \citet{Nokland2019}. The main motivation was to work with an architecture that allows pure layer-wise training which is impossible in e.g. ResNet-50 due to skip-connections.
Surprisingly we find that the transition from ResNet-50 to VGG-6 does neither compromise CPC losses nor downstream classification performance for almost all training methods, see \autoref{tab:arch}.

\begin{table}[h]
    \centering
    \begin{tabular}{cc}
    \hline
         \# of trainable layer & layer type \\\hline\hline
         1 & 3$\times$3 conv128, ReLU \\\hline
         2 & 3$\times$3 conv256, ReLU \\
           & 2$\times$2 MaxPool \\\hline
         3 & 3$\times$3 conv256, ReLU \\\hline
         4 & 3$\times$3 conv512, ReLU \\
           &  2$\times$2 MaxPool \\\hline
         5 & 3$\times$3 conv1024, ReLU \\
           & 2$\times$2 MaxPool \\\hline
         6 & 3$\times$3 conv1024, ReLU \\
           & 2$\times$2 MaxPool \\
         \hline
    \end{tabular}
    \vspace{0.5cm}
    \caption{Architecture of the VGG-6 encoder network. Convolutional layers (conv) have stride (1, 1), Pooling layers use stride (2, 2). The architecture is inspired by the VGG-like networks used in \citet{Nokland2019}.}
    \label{tab:vgg6-arch}
\end{table}

In GIM and CLAPP, the encoder is split into several, gradient-isolated modules. Depending on the number of such modules, each module contains a different number of layers. 
In CPC we do not use any gradient blocking and consequently the encoder consists only of one module containing layers 1-6. 
In layer-wise GIM and CLAPP each of the 6 modules contains exactly on layer (and potentially another MaxPooling layer).
\autoref{tab:modules} shows the distribution of layers into modules for the cases in between.

\begin{table}[ht]
    \centering
    \caption{Distribution of layers into modules as done for the simulations in \autoref{tab:comp} and \autoref{tab:block}. The layer numbers refer to \autoref{tab:vgg6-arch}.}
    \begin{tabular}{cc}
        \hline
        \# of modules & layer distribution \\\hline\hline
        1 (CPC) & (1,2,3,4,5,6)\\
        2 & (1,2,3), (4,5,6)\\
        3 & (1,2), (3,4), (5,6)\\
        4 & (1,2,3),(4),(5),(6) or (1),(2),(3),(4,5,6)\\
        6 & (1),(2),(3),(4),(5),(6)\\
        \hline
    \end{tabular}
    
    \label{tab:modules}
\end{table}

\begin{table}[ht]
\centering
\caption{Linear classification test accuracy (\%) on STL-10 with features coming from two different encoder models: ResNet-50 as in \citet{Lowe2019} and a 6-layer VGG-like encoder (VGG-6). Values for ResNet-50 are taken from \citet{Lowe2019}. \label{tab:arch}}
\vspace{5pt}
\begin{tabular}{lcc} 
 \hline
  & ResNet-50 & VGG-6\\\hline\hline
 Random init & 27.0 & 21.8\\
 Greedy Supervised & 65.2 & 65.0\\
 Supervised & 71.4 & 73.2\\\hline
 CPC & 80.5 & 81.1\\
 GIM (3 modules) & \textbf{81.9} & 78.3 \\\hline
\end{tabular}
\end{table}


\paragraph{Reference algorithms}
{\it Random init} refers to the random initialisation of the encoder network. It thus represents an untrained network with random weight matrices. This `method' serves as a lower bound on performance and as a sanity check for other algorithms.

In classic {\it supervised} training, we add a fully-connected layer with as many output dimensions as classes in the data set to the encoder architecture. Then the whole stack is trained end-to-end using a standard supervised loss and back-propagation. For data sets offering many labels this serves as an upper bound on performance of unsupervised methods. In the case of sparsely labeled data, unsupervised learners could, or even should, outperform supervised learning.

The {\it greedy supervised} method trains every gradient-isolated module of the encoder separately. For that, one fully-connected layer is added on top of each module. Then, for every module, the stack consisting of the module and the added fully-connected layer is trained with a standard supervised loss requiring labels.
Gradients are back-propagated within the module but blocked between modules. 
This layer-wise training makes the method quasi layer-local, however, BP through the added fully-connected layer is still required.

\subsection{Audio experiments}

We follow most of the implementation methods used in \citet{Lowe2019}.
The model is trained without supervision on 100 hours of clean spoken sentences from the LibriSpeech data set \citep{Panayotov2015} without any data augmentation.
For feature evaluation, a linear classifier is used to extract the phonemes divided into 41 classes. This classifier is trained on the test split of the same dataset, along with the phoneme annotations computed with a software from \citet{Oord2018}. 

The audio stream is first processed with four 1D convolutional layers and one recurrent layer of Gated Recurrent Units (GRU). The hyperparameters of this architecture are the same as the ones used in \citet{Lowe2019}.

All convolutional layers are assigned a CPC or a CLAPP loss as described in the main text and the gradients are blocked between them.
To train the last layer -- the recurrent layer --, we add one variant of the CLAPP and CPC losses where the score function is defined by $u_{t}^{\tau} = \bm {z^{\tau} } {}^{\top} \bm W^{\mathrm{pred}} \bm c^t$ where $\bm c^t$ is the activity of the GRU layer and $\bm z^{\tau}$ is the activity of the last layer of convolutions.
This loss is minimized with respect to the parameters of $\bm c$ and $\bm z$, and the gradients cannot flow to the layers below (hence $\nabla_{\bm W^{\bm z}} \bm c^t = \bm 0$ even if $\bm z$ is implicitly the input to $\bm c$ with this architecture).

Within the GRU layer the usual implementation of gradient descent with pytorch involves back-propagation through time (BPTT), even if we avoided BP between layers. To avoid all usage of back-propagation and obtain a more plausible learning rule we used e-prop \citep{bellecsolution} instead of BPTT. The details of this implementation are provided in the next section (\autoref{app:extra}) in the paragraph `Combining e-prop and CLAPP'.

\subsection{Video experiments}

\paragraph{General procedure} We use the UCF-101 dataset \citep{soomro2012ucf101}, an action recognition dataset containing 13,000 videos representing 101 actions. The original clips have a frequency of 30 frames per second and were downsampled by a factor 3. Videos were cut into clips of respectively 54 frames (5.4 seconds) for self-supervised learning and 72 (7.2 seconds) for the following classification. Frames in a clip were randomly grayed and jittered following the procedure of \citet{Han2019}. Cropping and horizontal flipping were applied per clip.

\paragraph{Architecture and training} For our network, we use a VGG-like network with 5 trainable layers presented in table \ref{tab:vgg5-arch}. The architecture is decomposed into spatial convolutions processing frames individually and additional temporal convolutions accounting for the temporal component of a clip. The first convolution uses no padding and all others have padding (0, 1, 1). The stride used for the spatial convolutions is, respectively, (1, 2, 2), (1, 2, 2) and (1, 1, 1) whereas the temporal convolutions both have stride (3, 1, 1) to prevent temporal overlap between successive encodings. 

\begin{table}[h]
    \centering
    \caption{Architecture of the VGG-5 encoder network. Pooling layers use stride (1, 2, 2).}
    \vspace{0.5cm}
    \begin{tabular}{cc}
    \hline
         \# of trainable layer & layer type \\\hline\hline
         1 & 1$\times$7$\times$7 conv96, BN, ReLU \\
           & 1$\times$3$\times$3 MaxPool \\\hline
         2 & 1$\times$5$\times$5 conv256, BN, ReLU \\
           & 1$\times$3$\times$3 MaxPool \\\hline
         3 & 1$\times$3$\times$3 conv512, BN, ReLU \\\hline
         4 & 3$\times$3$\times$3 conv512, BN, ReLU \\\hline
         5 & 3$\times$3$\times$3 conv512, BN, ReLU \\
           & 1$\times$3$\times$3 MaxPool \\
         \hline
    \end{tabular}
    \vspace{0.5cm}
    \label{tab:vgg5-arch}
\end{table}

Whereas \cite{Lowe2019} applies pooling to the feature maps outputted by a layer to obtain the encoding, we flatten them to preserve spatial information necessary to understand and predict the spatial flow and structure from movements related to an action.

For the training procedure, we use a batch size of 8 and train for 300 epochs with a fixed learning rate of 0.001. We use as many negative samples as available in the batch, for the spatial convolutions this leads to 429 negatives and the two temporal convolutions respectively have 141 and 45. This decrease is due to the temporal reductions occurring, aimed at preventing information leakage between sequences. 

\section{Additional material}\label{app:extra}

\paragraph{Weight transport in $\bm W^{\mathrm{pred}}$}
The update of the encoder weights ${\bf W}$ in CPC, GIM and CLAPP (before introducing $\bm W^{\mathrm{retro}}$) relies on weight transport in $W^{\mathrm{pred}}$, as seen in \autoref{eq:grad_w_c} or \autoref{eq:grad-clapp-2}.

The activity of $\bm c^t$ is propagated with the matrix $\bm W^{\mathrm{pred}}$ and $\bm z^{\tau}$ with its transpose. This is problematic because typical synapses in the brain transmit information only in a single direction. The existence of a symmetric reverse connection matrix would solve this problem but raises the issue that connection strengths would have to be synchronised (hence the word {\it weight transport}) between $\bm W^{\mathrm{pred}}$ and the reverse connections.

One first naive solution is to block the gradient at the layer $\bm c$ in the definition of the score $u_{t}^{\tau} = \bm z^{\tau} {}^\top \bm W^{\mathrm{pred}} \mathrm{block\_grad}(\bm c^{t})$, with the definition:
\begin{eqnarray}\nonumber
    \mathrm{block\_grad}(x) &=& x\\
    \nabla_x \mathrm{block\_grad}(x) &=& 0 ~.
    \label{eq:gradientblockfunction}
\end{eqnarray}
In this way, no information needs to be transmitted through the transpose of $\bm W^{\mathrm{pred}}$. However this results in a relatively large drop in performance on STL-10 for Hinge Loss CPC (78.0 \%) and CLAPP (70 \%).

A better option -- and as done in the main paper -- is to split the original $\bm W^{\mathrm{pred}}_{\mathrm{orig}}$ into two matrices $\bm W^{\mathrm{pred}}$ and  $\bm W^{\mathrm{retro}}$ (for `retrodiction') which are independent and which allow information flow only in a single direction (as in actual biological synapses).
To this end, we split the loss function into two parts: one part receives the activity $\bm W^{\mathrm{pred}} \bm c^t$ coming from $\bm c^t$ and only updates the parameters of $\bm z$; and the other part receives the activity $\bm W^{\mathrm{retro}} \bm z^{\tau}$ coming from $\bm z^{\tau}$ and updates the parameters of $\bm c$. 
Like this information is transmitted through $\bm W^{\mathrm{pred}}$ and $\bm W^{\mathrm{retro}}$ instead of $\bm W^{\mathrm{pred}}$ and its transpose matrix and hence solves the weight transport problem.

More formally, let us write $F$ to summarize the definition of the usual CLAPP loss function in \autoref{eq:CLAPP_loss} such that $\mathcal{L}_{\mathrm{CLAPP}}^{t}=F(\bm c^t, \bm z^{\tau}, \bm W^{\mathrm{pred}}_{\mathrm{orig}})$. We then introduce a modified version of the CLAPP loss function:
\begin{equation}
    \Tilde{\mathcal{L}}_{\mathrm{CLAPP}}^{t}  =
    \frac{1}{2} \left( \Tilde{\mathcal{L}}_{\mathrm{CLAPP}}^{t,\bm z} 
    +
    \Tilde{\mathcal{L}}_{\mathrm{CLAPP}}^{t,\bm c} \right)~,
    \label{eq:CLAPP_loss_tilde}
\end{equation}
with $\Tilde{\mathcal{L}}_{\mathrm{CLAPP}}^{t,\bm z} = F(\mathrm{block\_grad}(\bm c^t), \bm z^{\tau}, \bm W^{\mathrm{pred}})$ and $\mathcal{L}_{\mathrm{CLAPP}}^{t, \bm c} = F(\bm c^t, \mathrm{block\_grad}(\bm z^{\tau}), \bm W^{\mathrm{retro}})$.
Similarly, we define the corresponding scores as $u_t^{\tau, \bm z} = \bm z^{\tau} {}^\top \bm W^{\mathrm{pred}} \mathrm{block\_grad}(\bm c^{t})$
and  $u_t^{\tau, \bm c} = \mathrm{block\_grad}(\bm z^{\tau}) {}^\top \bm W^{\mathrm{retro}, \top} \bm c^{t}$. 
With this, the gradients with respect to the weight parameters $W^{\bm z}_{ji}$ (encoding $\bm z^{\tau}$) are:
\begin{equation}
    \frac{\partial u_{t}^{\tau,\bm z}}{\partial W^{\bm z}_{ji}}  = 
     x_{i}^{\tau, \bm z} \rho'(a_j^{\tau,\bm z})
     (\bm W^{\mathrm{pred}} \bm c^{t})_j
     \hspace{1cm} \text{and} \hspace{1cm}
     \frac{\partial {u}_{t}^{\tau,\bm c}}{\partial W^{\bm z}_{ji}} = 0~,
     \label{eq:weight_transport_1}
\end{equation}
and the gradients with respect to the weights $W^{\bm c}_{ji}$ (encoding $\bm c^t$) become:
\begin{equation}
\frac{\partial {u}_{t}^{\tau,\bm z}}{\partial W^{\bm c}_{ji}} = 0
     \hspace{1cm} \text{and} \hspace{1cm}
    \frac{\partial u_{t}^{\tau, \bm c}}{\partial W^{\bm c}_{ji}}  =
     x_{i}^{t, \bm c} \rho'(a_j^{t,\bm c})
     (\bm W^{\mathrm{retro}} \bm z^{\tau})_j~.
     \label{eq:weight_transport_1}
\end{equation}


The final plasticity rule combines those terms and recovers the original CLAPP rule \autoref{eq:dw_z_CLAPP} and \autoref{eq:dw_c_CLAPP}:

\begin{eqnarray}\nonumber
    \Delta W_{ji}^{\tau} &=& \gamma_{\tau}\left[\Delta W_{ji}^{\tau, \bm z} + \Delta W_{ji}^{\tau, \bm c}\right] \\\nonumber
    \Delta W_{ji}^{\tau, \bm z} &=& 
    \left(\bm W^{\mathrm{pred}} \bm c^t\right)_j ~ \rho'(a_j^{\tau, \bm z}) ~ x^{\tau, \bm z}_i \\
    \Delta W_{ji}^{\tau, \bm c} &=&
    \left(\bm W^{\mathrm{retro}} \bm z^{\tau}\right)_j ~ \rho'(a_j^{t, \bm c}) ~ x^{t, \bm c}_i,
\end{eqnarray}
under the assumption of having only one gating factor $\gamma_{\tau}$. This is approximately the case when $\bm W^{\mathrm{pred}}$ and $\bm W^{\mathrm{retro}}$ align since then $u_{t}^{\tau} = u_{t}^{\tau, \bm c} = u_{t}^{\tau, \bm z}$. We consider this assumption realistic since $\bm W^{\mathrm{pred}}$ and $\bm W^{\mathrm{retro}}$ share the same update rule \autoref{eq:dw_pred_CLAPP}.
We see that the propagation of the activity through the independent weights $\bm W^{\mathrm{pred}}$ and $\bm W^{\mathrm{retro}}$ is always unidirectional.

It turns out that, using the modified loss $\Tilde{\mathcal{L}}_{\mathrm{CLAPP}}^{t}$, \autoref{eq:CLAPP_loss_tilde}, instead of the original CLAPP loss $\mathcal{L}_{\mathrm{CLAPP}}^{t}$, \autoref{eq:CLAPP_loss}, the performance on STL-10 (linear classification on last layer) is unchanged for Hinge Loss CPC (80.2 \%) and CLAPP-s (74.1 \%).

\paragraph{Combining e-prop and CLAPP}
CLAPP avoids the usage of back-propagation through the depth of the network, but when using a recurrent GRU layer in the audio task, gradients are still back-propagated through time inside the layer.
A more plausible alternative algorithm has been suggested in \citet{bellecsolution}: synaptic eligibility traces compute local gradients forward in time using the activity of pre- and post-synaptic units, then these traces are merged with the learning signal (here 
$\bm W \bm z^{t + \delta t}$) to form the weight update.
It is simple to implement e-prop with an auto-differentiation software such as pytorch by introducing a $\mathrm{block\_grad}$ function in the update of the recurrent network. With GRU, we implement a custom recurrent network as follows (the notations are consistent with the pytorch tutorial on GRU networks\footnote{\url{https://pytorch.org/docs/stable/generated/torch.nn.GRU.html}} and unrelated to the rest of the paper):
\begin{eqnarray}
\bm r_t & = & \sigma (\bm W_{ir} \bm x_t + \bm b_{ir} + \bm W_{hr}  \mathrm{block\_grad}( \bm h_{t-1} ) + \bm b_{hr}) \\
\bm z_t & = & \sigma( \bm W_{iz} \bm x_t + \bm b_{iz} + \bm W_{hz} \mathrm{block\_grad} ( \bm h_{t-1} ) +\bm b_{hz}) \\
\bm n_t & = & \operatorname{tanh} \left( \bm W_{in} \bm x_t + \bm b_{in} + \bm r_t \star (\bm W_{hn} \mathrm{block\_grad} ( \bm h_{t-1} )+ \bm b_{hn})\right) \\
\bm h_t & = & (1- \bm z_t) \star \bm n_t + \bm z_{t} \star \bm h_{t-1}
\end{eqnarray}
In summary we use $\bm h_t$ as the hidden state of the recurrent network, $\bm r_t$, $\bm z_t$ and $\bm n_t$ as the network gates, $\star$ as the term-by-term product, and $\bm W_{\cdot}$ and $\bm b_{\cdot}$ as the weights and bias respectively.
One can show that applying e-prop in a classical GRU network is mathematically equivalent to applying BPTT in the network above.

In simulations, we evaluate the performance as the phoneme classification accuracy on the test set. We find that CLAPP-s achieves $61.7\%$ with BPTT and $58.6\%$ with e-prop; but the latter can be implemented with purely local learning rules by relying on eligibility traces \citep{bellecsolution}. In comparison, phoneme classification from the last feedforward layer before the RNN only yields $52.4\%$ accuracy.

\paragraph{Biologically plausible computation of the score $u_t^{t+\delta t}$}
We think of the loss $\mathcal{L}_{CLAPP}^{t}$ in \autoref{eq:CLAPP_loss} as a surprise signal that is positive if the prediction is wrong, either because a fixation has been wrongly predicted as a saccade or vice-versa. Surprising events are indicated by physiological markers of brain activity such as the EEG or pupil dilation.
Moreover, the activity of neuromodulators such as norepinephrine, acetylcholine, and partially also dopamine is correlated with surprising events; an active sub-field of computational neuroscience attempts to relate neuro-modulators to surprise and uncertainty \citep{angela2005uncertainty, nassar2012rational, ostwald2012evidence, schwartenbeck2013exploration, heilbron2019confidence, liakoni2021learning}. 

In analogy to the  theory of reinforcement learning, where abstract models have been successfully correlated with brain activity and dopamine signals well before the precise brain circuitry necessary to calculate the dopamine signal was known \citep{dabney2020distributional}, we take the view that surprise signals exist and can be used in the models, even if we have not yet identified a circuit to calculate them. The neuromodulator signal in our model would be 1 if $\mathcal{L}_{CLAPP}^{t} > 0$ and zero otherwise. Thus the exact value of $\mathcal{L}_{CLAPP}^{t}$ is not needed.

Nevertheless, let us try to sketch a mechanism to compute this signal. Every neuron $i$ has access to its `own' internal dendritic signal $\hat{z}_i^t = \sum_j W^{pred}_{ij} c^t_j$ interpretable as the dendritic prediction of somatic activity \citep{urbanczik2014learning}. 
What we need is the product $z_i^{t+\delta t} \, \hat{z_i}^t$ and then we need to sum over all neurons. Four insights are important.
First, a potential problem is that the dendritic prediction $\hat{z}_i^t$ is \emph{different} from the actual activity $z_i^{t+\delta t}$ that is driven in our model by feedforward input. However, the work of \citet{Larkum1999} has shown that neurons emit specific burst-like signals if both dendrite and soma are activated. The product $z_i^{t+\delta t} \, \hat{z_i}^t$ can be seen as a detector of such coincident events.  Second, if bursts indicate such coincident events, then the burst signals of many neurons need to be summed together, which could be done either by an interneuron in the same area (same layer of the model) or by neurons in a deep nucleus located below the cortex. The activity of this nucleus would serve as one of the inputs of the nucleus that actually calculates surprise.
Third, following ideas on time-multiplexing in \citet{payeur2021burst}, the burst signal can be considered as a communication channel that is {\em separate} from the single-spike communication channel for the feedforward network used for inference. 
Fourth, as often in neuroscience, positive and negative signals must be treated in different pathways (the standard example is ON and OFF cells in the visual system), before they would be finally combined with the saccade signal $y$ to emit  the binary surprise signal $\gamma_t=y^t H^t$ that is broadcasted to the  area corresponding to one layer of our network. 
An empirical test of this suggested circuitry is out of scope for the present paper but will be addressed in future work.

\begin{figure}[ht]
    \centering
    \includegraphics[width=0.35\textwidth]{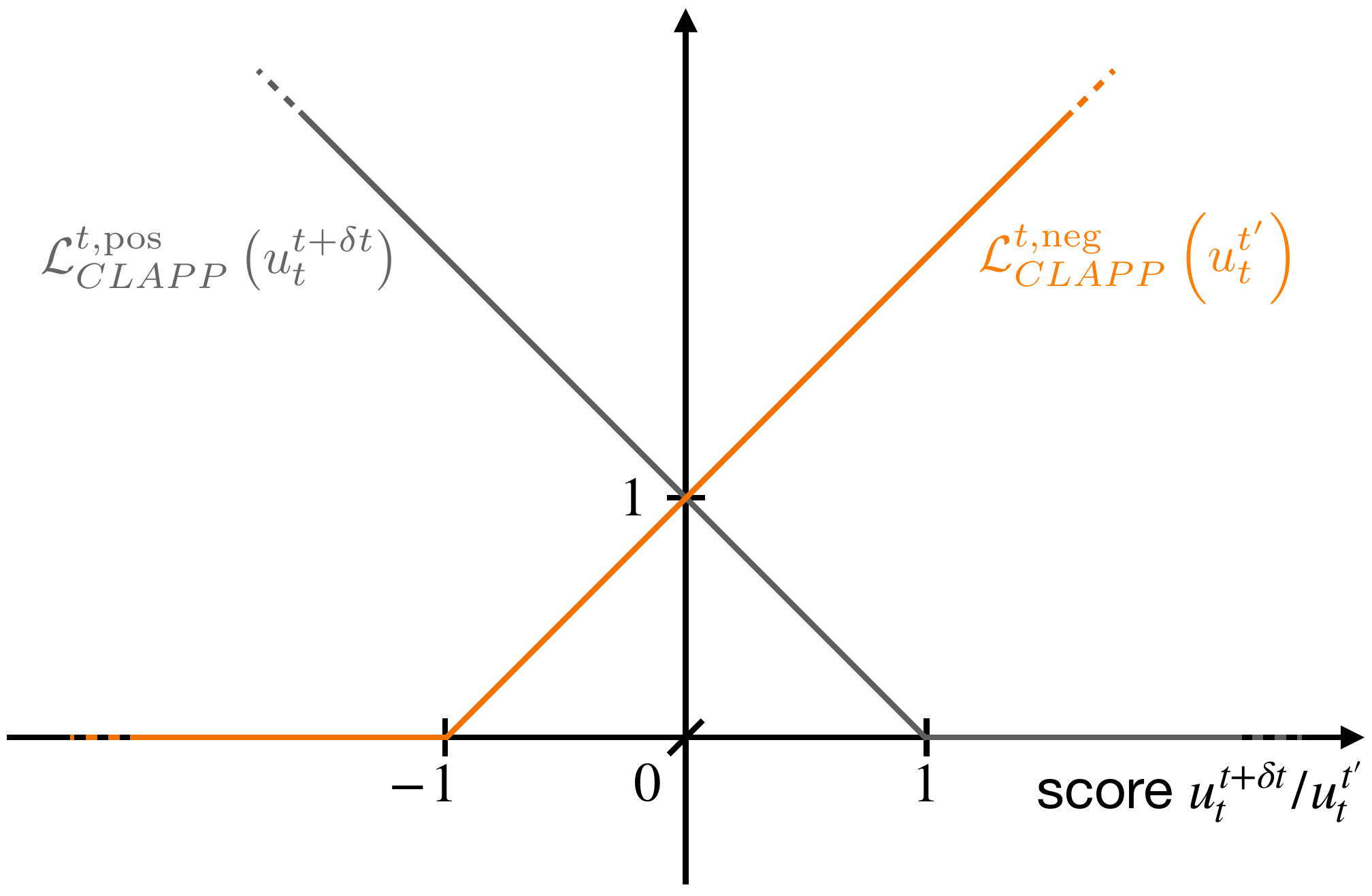}
    \caption{Illustration of the positive ($y=+1$, gray) and negative ($y=-1$, orange) part of the CLAPP loss, see \autoref{eq:CLAPP_loss}}
    \label{fig:CLAPP_loss}
\end{figure}

\begin{figure}[ht]
    \centering
    \includegraphics[width=0.8\textwidth]{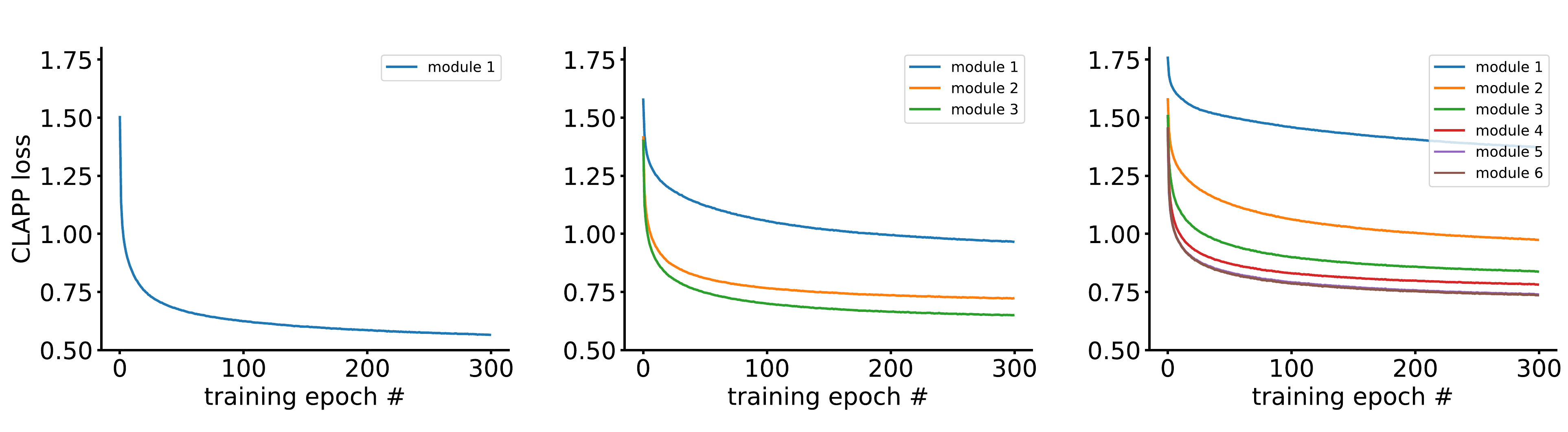}
    \caption{CLAPP-s training losses for encoders split into 1, 3 or 6 gradient-isolated modules.}
    \label{fig:CLAPP_loss_curves}
\end{figure}

\begin{table}[ht]
\centering
\caption{Linear classification test accuracy (\%) on STL-10 with features from a VGG-6 encoder trained with CLAPP-s for different sizes of gradient-isolated modules. \label{tab:block}}
\vspace{5pt}
\begin{tabular}{lcc} 
 \hline
 \# modules & \# layers per module & Test accuracy (\%)\\\hline\hline
 6, i.e. layer-wise (CLAPP) & 1 & 74.0\\\hline
 4 modules upper & 3,1,1,1 & 75.4 \\
 4 modules lower & 1,1,1,3 & 76.2 \\
 3 modules & 2 & 77.4 \\
 2 modules & 3 & 77.6 \\\hline
 1 module (end-to-end) (see \autoref{tab:comp}) & 6 & 80.3 \\\hline
\end{tabular}
\end{table}

\clearpage

\begin{figure}[h]
    \centering
    \includegraphics[width=0.9\textwidth]{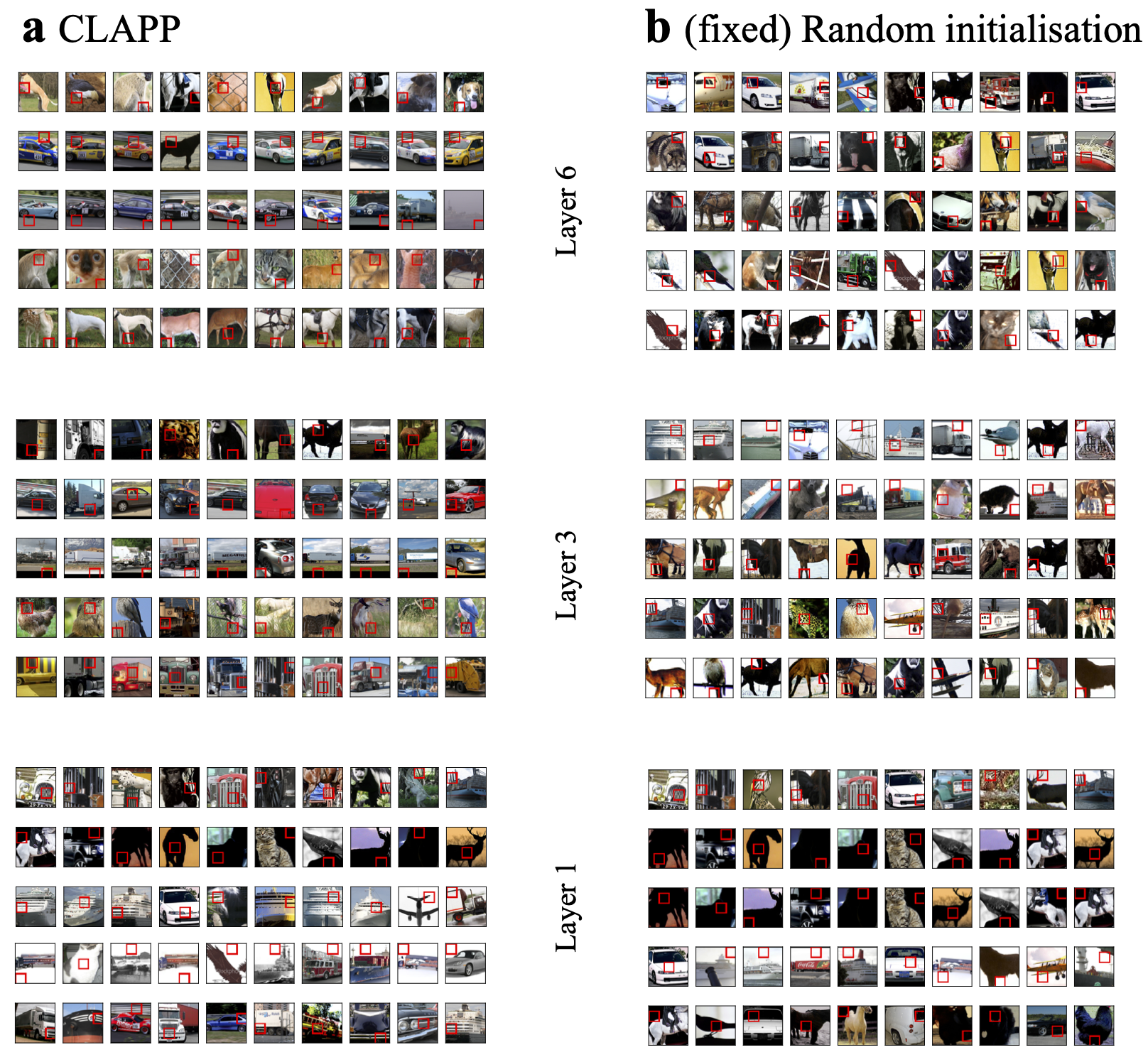}
    \caption{(As \autoref{fig:vis} a) Red boxes in STL-10 images indicate patches that best activate a specific neuron (rows) in \textbf{a} a network trained with CLAPP or \textbf{b} a random encoder with random weights fixed at initialisation.
    For CLAPP, layer 1 extracts simple features like gratings or uniform patches, whereas higher layers extract richer features like object parts.
    For the random encoder, no clear hierarchy is apparent across layers.}
    \label{fig:patch_vis_random}
\end{figure}

\paragraph{Preferred patch visualisation for random encoder}
As a control, we repeat the preferred patch visualisation analysis, as in \autoref{fig:vis} a, for the random encoder, i.e. a network with random fixed weights.
The result is shown in \autoref{fig:patch_vis_random} b, in comparison with the analysis of an encoder trained with CLAPP.
For CLAPP, higher layers extract higher-level features creating a hierarchy, whereas for the random encoder, no clear hierarchy is apparent across layers.
Together with the non-informative t-SNE embedding of the random encoder (\autoref{fig:vis} b), this suggest that a convolutional architecture alone does not yield hierarchical representations.
    
\paragraph{Gradient flow through MaxPooling layers}
In fact, MaxPooling can be viewed as a simple model of lateral inhibition which provides a learning rule compatible with \autoref{eq:rule_prototype}, without introducing approximations and without blocking gradients below the MaxPool operator.

The idea is that $2 \times 2$ MaxPooling can be viewed as a simple model of lateral inhibition between the 4 neurons involved. 
During inference, this inhibition enforces only one of the four neurons to be active. 
We use the following notation for the output of the MaxPool operator ${z'}_{j'}^t = \max \{ {z}_{j_0}^t, {z}_{j_1}^t, {z}_{j_2}^t, {z}_{j_3}^t \}$, where ${z}_{i}^t$ is defined as in the main paper.
We define ${\bm c'}^{t}$ accordingly for the context layer, if it includes a MaxPool operator.

Then, following the derivation from the main text, the learning rule is proportional to the gradient $\frac{\partial u_t^{t+\delta t}}{\partial W_{ij}}$,
but now $u_t^{t+\delta t}$ is defined using the output of the pooling operators:
$u_t^{t+\delta t} = \sum_{k',j'} {z'}^{t+\delta t}_{j'} W^{\mathrm{pred}}_{j'k'} {c'}^{t}_{k'}$. 
Since the partial derivative over the MaxPool operator is either $1$ (the neurons is active), or $0$ (for the other three neurons, which are inhibited), $\frac{\partial u_t^{t+\delta t}}{\partial W_{ij}}$ is either $\left(\sum_{k'} W^{\mathrm{pred}}_{j'k'} {c'}^{t}_{k'}\right) \cdot \sigma'(a_{j'}^{t+\delta t}) x_i^{t+\delta t}$ if $j = j'$ (the neuron is active), or $0$ if $j \neq j'$ (the neuron is inhibited).
Hence, and without further approximation, the learning rule is only applied if the neuron is active, in which case $\frac{\partial u_t^{t+\delta t}}{\partial W_{ij}}$ takes the form `dendritic signal $\times$ post $\times$ pre', and the resulting learning rule is compatible with \autoref{eq:rule_prototype}.

\paragraph{Predicting from higher layers}
We ran CLAPP-s with the context representation $c^t$ coming from one layer above the predicted layer $z^{t+\delta t}$ (except for the last layer, where $c^t$ and $z^{t+\delta t}$ come from the same layer).
Linear classification performance on STL-10 still grows over layers but only yields 72.4 \% test accuracy when classifying from the last layer.
In comparison defining $\bm c^t$ to be the same layer as $\bm z^t$ reached 75.0\%.

\end{document}